\pdfoutput=1
\documentclass[11pt]{article}

\usepackage{acl}

\usepackage{times}
\usepackage{latexsym}
\usepackage{subfigure}
\usepackage{booktabs} 
\usepackage{graphicx}
\usepackage{amsmath}
\usepackage{amssymb}
\usepackage{mathtools}
\usepackage{amsthm}
\usepackage{graphicx}
\usepackage{multirow}
\usepackage{xspace}
\usepackage{listings}
\usepackage{xcolor}

\usepackage[T1]{fontenc}
\usepackage[utf8]{inputenc}

\usepackage{microtype}

\newcommand{\ours}{\textsc{LVChat}\xspace}

\usepackage{amsmath,amsfonts,bm,amssymb,mathtools,amsthm}
\usepackage{color,xcolor,xspace}
\usepackage{booktabs}
\usepackage{thm-restate}

\def\eqref#1{Eq.~(\ref{#1})}

\def\1{\bm{1}}

\def\rmE{{\mathbf{E}}}

\def\rmI{{\mathbf{I}}}

\def\rmP{{\mathbf{P}}}

\def\rmV{{\mathbf{V}}}
\def\rmW{{\mathbf{W}}}

\DeclareMathAlphabet{\mathsfit}{\encodingdefault}{\sfdefault}{m}{sl}
\SetMathAlphabet{\mathsfit}{bold}{\encodingdefault}{\sfdefault}{bx}{n}

\newcommand{\duration}{T} %
\newcommand{\totalframes}{F} %
\newcommand{\clipframes}{K} %
\newcommand{\cliptokens}{N} %
\newcommand{\sampledframes}{F_s} %
\newcommand{\numofclips}{n} %
\newcommand{\maxnumofclips}{n_m} %
\newcommand{\numofclipsinter}{n_i} %
\newcommand{\numinter}{\gamma}

\title{\ours: Facilitating Long Video Comprehension}

\author{Yu Wang$^*$ \\
  UC San Diego \\
  \texttt{yuw164@ucsd.edu} \\\And
  Zeyuan Zhang\thanks{$\,\,\,$Equal Contribution.} \\
  UC San Diego \\
  \texttt{zez018@ucsd.edu}  \\\And
  Julian McAuley\\
  UC San Diego\\
  \texttt{jmcauley@ucsd.edu} \\\And
  Zexue He \\
  UC San Diego\\
  \texttt{zehe@ucsd.edu}
}

\begin{document}
\maketitle
\begin{abstract}

Enabling large language models (LLMs) to read videos is vital for multimodal LLMs. Existing works show promise on short videos whereas long video (longer than e.g.~1 minute) comprehension remains challenging. The major problem lies in the over-compression of videos, i.e., the encoded video representations  are not enough to represent the whole video. To address this issue, we propose Long Video Chat (\ours), 
where Frame-Scalable Encoding (FSE) is introduced to dynamically adjust the number of embeddings in alignment with the duration of the video to ensure long videos are not overly compressed into a few embeddings. 
%However, as video duration extends, the resultant increase in tokens may exceed the input window of capacities. Consequently, we propose Interleaved Frame Encoding (IFE) to facilitate the management of embeddings.
To deal with long videos whose length is beyond videos seen during training,  we propose Interleaved Frame Encoding (IFE), repeating positional embedding and interleaving multiple groups of videos to enable long video input, avoiding performance degradation due to overly long videos. 
% , a novel encoding strategy of long input whose global dependency is captured into reorganized short clips which are slightly shifted along with time, 
% to avoid performance degradation due to the out-of-distribution issues.
Experimental results show that \ours significantly outperforms 
%all the 
existing methods by up to 27\% in accuracy on long-video QA datasets and long-video captioning benchmarks. Our code is published at \url{https://github.com/wangyu-ustc/LVChat}. 

%Enabling large language models (LLMs) to read videos is vital for multimodal LLMs. Existing works can read videos whereas long video (typically longer than 1 minute) comprehension remains challenging. The major problem lies in the over-compression of videos, i.e., the mapped video representations in the word embedding domain are not enough to represent the whole video. To address this issue, we propose Long Video Chat (\ours). At its core, we introduce Frame-Scalable Encoding (FSE) to dynamically adjust the volume of embeddings in alignment with the duration of the video to ensure long videos are not overly compressed into a few embeddings. However, as video duration extends, the resultant increase in tokens may exceed processing capacities. Consequently, we propose Interleaved Frame Encoding (IFE) to facilitate the management of embeddings. Experimental results show that LV-Chat could outperform all the existing methods on both the synthetic datasets (extend short video QA into long video QA) and the real-world dataset (captioning tasks on long videos).
\end{abstract}

\section{Introduction}
\label{introduction}

Recent works have been proposed to enhance  the multimodal capabilities of large language models, extending their power beyond text to  other data modalities such as images \cite{touvron2021training,bao2021beit, he2022masked} and audio \cite{hassid2023textually,borsos2023audiolm,sicherman2023analysing}. Among them, videos offer a unique medium through how humans perceive the real world \cite{li_videochat}. To leverage this, recent efforts on augmenting LLMs' video comprehension have focused on finetuning LLMs with video instruction data such as VideoChat \cite{li_videochat}, VideoChatGPT \cite{video-chatgpt}, VideoLlama \cite{videollama}.

\begin{figure}
    \centering
    \includegraphics[width=\linewidth]{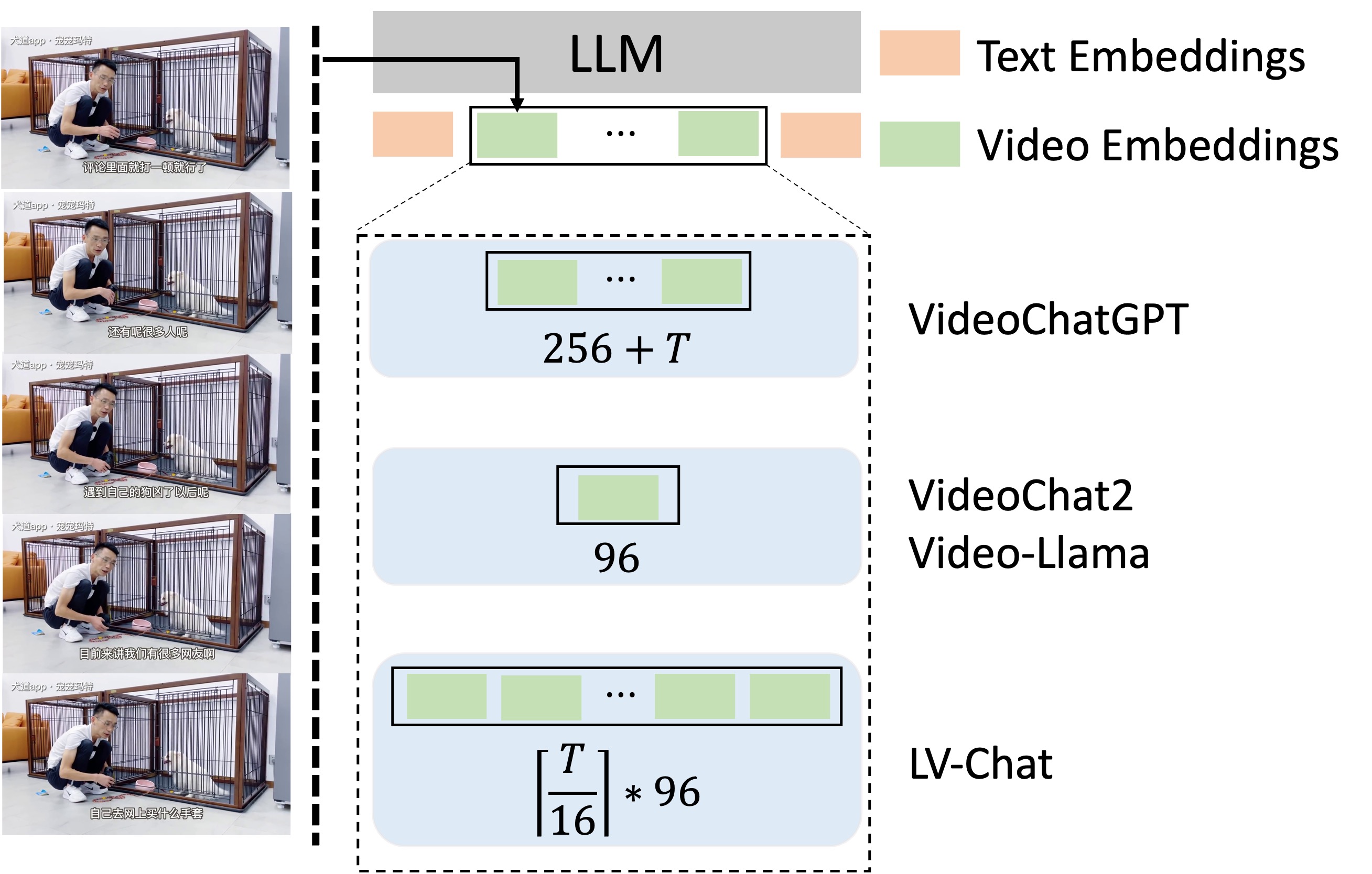}
    \caption{Previous video language models may suffer from over-compression for long video modeling (e.g., $T>60$s ) since a limited number of video tokens are used in LMs. In contrast, \ours 
 demonstrates superior performance on long videos by modeling more video tokens. }
    \label{fig:teaser}
\end{figure}

%
%Though previous video-language models show promising results (especially on short videos), modeling videos longer than 1 minute is shown to be challenging in  these works \cite{}. We hypothesize the reason that \emph{the inability to comprehend long videos comes from the over-compression of the videos.} For instance, with a video of $\duration$ seconds ($\duration$ can be up to 600), VideoChatGPT~\citep{videochatgpt} would sample $\sampledframes$ and then compress the $\duration$-second video into $256 + \sampledframes$ tokens (256 is prefix tokens for global information), which is not enough when video gets long. Especially, VideoChat~\citep{li_videochat} and Video-Llama~\citep{video-llama} sample $\sampledframes$ frames and convert them into $96$ tokens, this means even $\duration$ is up to 600, the number of tokens may be overly tiny to represent the video. 

%Zexue's
While previous video-language models have demonstrated promising results, particularly with short videos, their performance on videos \textit{longer than one-minute} is observed to be challenging \citep{li_videochat}.  We believe (and empirically prove it in our experiments) that \emph{the inability to comprehend long videos comes from the over-compression of video content.} 
For example, VideoChatGPT~\cite{video-chatgpt} models a video of $\duration$ seconds 
% (where $\duration$ can extend up to 600 seconds) 
by sampling ($\totalframes$ frames). These frames, along with a prefix of 256 tokens designated for global information, are then compressed into a total of $256 + \totalframes$ tokens. This compression strategy is insufficient for longer videos, where the complexity and information density exceed the representational capacity of the allocated tokens. 
% What's more, 
On the other hand, mainly focusing on short videos,  
VideoChat~\cite{li_videochat} and Video-Llama~\cite{videollama}  
convert $\sampledframes$ sampled frames into a fixed tiny number of embeddings (96 embeddings), regardless of the video's duration,
% (even up to 600 seconds), 
resulting in inadequate information for effective long-video representation.

%yu's
%For instance, with a video of $\duration$ seconds ($\duration$ can be up to 600),  VideoChatGPT~\citep{videochatgpt} would sample $\sampledframes$ and then uses $256 + \sampledframes$ tokens to represent the video. Though $\sampledframes$ can be as large as several hundred, using $256 + \sampledframes$ to represent $\sampledframes$ does not seem to be enough. This intuition comes from (1) MiniGPT4 proposes to use 64 tokens to represent one single image, (2) \cite{} claims one image is worth a thousand tokens. Meanwhile, VideoChat~\citep{li_videochat} and Video-Llama~\citep{video-llama} sample $\sampledframes$ frames and convert them into $96$ tokens, this means even $\duration$ is up to 600, the number of tokens may be overly few to represent the video. 

In this work, we focus on the long video understanding scenario and propose a novel video language model \ours. \ours has two key components: \textit{Frame Scalable Encoding (FSE)} and \textit{Interleaved Frame Encoding (IFE)}. To tackle the over-compression problems, we design FSE, a new feature extraction strategy that scales the number of tokens with the video length $\duration$. Specifically, every 16 frames are compressed into 96 tokens to ensure the video information is mostly maintained during the mapping. The model is then fine-tuned on these compressed $\lceil\frac{T}{16}\rceil*96$ embeddings.  To overcome the out-of-distribution (OOD) problem encountered during inference when videos are longer than those seen during training, 
we introduce IFE, a novel interleaving strategy to repeat positional embeddings and interleave multiple groups of videos to enable long video input and avoid OOD issue. 

We evaluate  \ours in the tasks of long-video question answering (QA) and long-video captioning. We observe that existing video benchmarks \cite{mvbench} primarily annotate a short clip of the entire video where the ground truth label is located (with such annotation, previous works only input the clip instead of the entire video). Since such annotation requires human effort to locate the answer, in our work, we investigate a more practical setup where such timestamp annotation is not available. To this end, we develop a long-video QA benchmark by randomly concatenating real video segments in MVBench \cite{mvbench} with distractor videos, along with a long-video captioning dataset TACoS \cite{tacos} where we manually create the ground truth captions according to human-annotated subtitles for its long videos. We also test \ours on EgoSchema \cite{egoschema}, a challenging long-video QA benchmark. 
% whose video is XXX long. 
The experimental results show that \ours largely improves the accuracy over baselines in our curated long-video QA task (600s) even with FSE only (up to 21\% improvement in accuracy) and adding IFE further improves accuracy (up to 27\%), 
% while at the same time maintaining superior performance with shorter videos (100s), 
highlighting the potential of \ours and shedding light on future advancements in long-video language models.

%Noticing that previous related benchmarks are designed with  the ground truth label of the input being annotated within a short clip of the entire video (thus previous works only model the small clip instead of the entire video), instead, in this work we design a long-video QA benchmark where the key video are concatenated with distractor videos and two long-video caption generation datasets extended from the TACoS and EgoSchema. We then test \ours in this more realistic setup where such timestamp annotation is missing. Experimental results show that \ours only with FSE further gain up to 21\% accuracy on long video QA task (600s) compared to baselines and adding IFE can even improve the accuracy by 27\%, while still being superior on short video (100s), demonstrating the effectiveness of \ours. As the initial trial, we hope \ours shed light on future works in developing language models that are capbable of capturing long video dependency.
%Yu's
%Our contribution could be summarized as follows: (1) We propose the hypothesis that the inability of current methods on long video comprehension is due to the overly compression, which leads to our solutions. (2) We finetune a new model with FSE, which is shown to largely improve the model's ability on long videos. (3) We further finetune a model with IFE to further fascilate long video comprehension. (4) Experimental results demonstrate the supriority of our models on long videos while maintaining the comprehension ability on short videos. 
\section{Related Work}

\subsection{Long Context Modeling}
There are lots of long context modeling techniques, among which modifying positional embeddings resembles our method the most. The most similar one to Interleaved Frame Encoding is Self-Exntending LLMs~\citep{self-extend-LLM}. Other works include adopting relative positional embedding~\citep{alibi}, positional interpolation~\citep{pi} and positional extrapolation~\citep{lex-transformer}. As mainly focus on text domain, these works are orthogonal to our use cases with multimodality data. 
%the advancements in this literature would remain inspirational.

\subsection{Video Question Answering}
Video Question Answering (VideoQA) has been a popular task for evaluating the model's ability to understand videos. Typical works pretrain a video-text model and perform a successive fine-tuning on VideoQA~\citep{zellers2021merlot, bain2021frozen, miech2019howto100m, wang2022all, fu2021violet, zeng2022x, li2022align}. These works are focused specifically on video question answering, and large language models are not introduced here, which might limit the interpretation of the video content and the application of the model.

% Video Question Answering(VideoQA) aims to align the dynamic visual contents with the linguistic semantics of a question to yield the answer.
% The recent paradigm is to first pretrain the model on a vast amount of video-text paired data~\cite{zellers2021merlot, bain2021frozen, miech2019howto100m} and fine-tune it on VideoQA~\cite{wang2022all, fu2021violet,  yang2022zero, bain2021frozen, zeng2022x, li2022align}.
% Typical VideoQA benchmarks take two formats: multiple-choice~\cite{li2020hero, tvqa} and open-ended~\cite{yang2021just, jang2017tgif, xu2017video, yu2019activitynet}.
% In contrast to multiple-choice VideoQA where several answer options are provided for each question, the goal of open-ended VideoQA is to predict the answer without any candidate answers.
% While existing open-ended VideoQA models~\cite{lei2021less,wang2022all,li2020hero,fu2021violet,zellers2021merlot,le2020hierarchical,yang2022zero} are promising, they still show sub-optimal performance due to the common practice of open-ended VideoQA that converts the task to a classification with only frequent answer candidates.
% To alleviate such issues, we introduce a novel benchmark to incorporate open-vocabulary setting into the VideoQA model.

% Previous literature about multi-modal LLMs includes the following categories: 
\subsection{Enabling LLMs to Process Videos through Descriptive Textualization}
A foundational approach towards equipping LLMs with video understanding capabilities involves the extraction of information from each frame of the video, subsequently converting this data into a textual format for LLM processing. Notable implementations of this strategy include ChatVideo~\citep{chatvideo} and VideoChat-Text~\citep{li_videochat}. These methods are limited by their reliance on textual conversion, which might pose problems when there are scenes beyond text descriptions. 
% ChatVideo~\citep{chatvideo},
% which utilizes a suite of foundation models to distill video frame data into textual information, which could be inputted into ChatGPT for processing; 
% VideoChat-Text~\citep{li_videochat}, 
% which textualizes video streams to obtain descriptive elements such as clip captions and tags for each frame. These texts are then analyzed by an LLM. 
% Despite their innovative nature, these methods are limited by their reliance on textual conversion, deviating from the intuitive, direct visual observed in human cognition. Moreover, certain visual scenes that elude precise description pose a significant challenge to these techniques.

\subsection{Enabling LLMs to Process Videos via Adapters}
An emergent trend in recent research focuses on introducing adapters to bridge the gap between visual representations and the textual embedding space of LLMs. Some works take the first step on images such as VC-GPT~\citep{VC-GPT}, VisualGPT~\citep{VisualGPT}, Mini-GPT4~\citep{minigpt4} and LlaVa~\citep{llava}, which proposes the adapters to map the visual encoder outputs into the word embedding space, enabling direct processing of image data with LLMs. Based on these models with image understanding capabilities, 
VideoChat-Embed and VideoChat2~\citep{li_videochat} propose to encode the videos into the embeddings with an extra adapter, where the visual encoder and the adapter is trained using video instruction datasets. Similarly, VideoChatGPT~\citep{video-chatgpt} initializes from LlaVa and is trained on another comprehensive video instruction dataset. Video-Llama~\citep{videollama} add audio modality into the instruction finetuning, enabling LLM to both see and hear. 
FrozenBiLM~\citep{yang2022zero} adapts a pre-trained BiLM to multi-modal inputs and introduces a set of additional modules including adapters, which are trained on video-text data. 
These methods show promising results in terms of short video understanding, but they may struggle with managing long videos (typically longer than $1$min). Our work is built based on the backbone VideoChat2~\citep{li_videochat} but with additional fine-tuning and design of FSE and IFE, improving long video understanding.

\section{Method}
\subsection{Preliminary}
\label{sub:preliminary}
Following VideoChat2~\citep{li_videochat}, assume we are generating captions for a given video $\rmV = [\rmI_i]_{i=1,2,\cdots,\totalframes}$, where $\totalframes$ is the total frames of the video, with $\rmI_i$ being the $i$-th frame. Then we need to convert the video $\rmV$ into embeddings $\rmE$: 
\begin{equation}
    \label{eq:video2embedding}
    \rmE = f_{vid} (\rmV).
\end{equation}
Here $f_{vid}$ denotes the video encoding model. Since we aim to enable the LLM to understand the video, $\rmE$ is usually trained to align the distribution of word embeddings in the LLMs $f_{llm}$. The next word is predicted following the equation:
\begin{equation}
    \label{eq:llm}
    \rmP = f_{llm} (\rmE, \rmW_{\leq t}),
\end{equation} 
where $\rmW_{\leq t}$ is the word embeddings of previous words generated in the sentence and $\rmP$ is the next-word probability distribution over the vocabulary. 
For the instantiation of the visual encoder $f_{vid}$ and the large language model $f_{llm}$, we follow VideoChat2~\citep{li_videochat}, where
UMT-L~\citep{umt} followed a pretrained QFormer and an extra linear adapter is used as $f_{vid}$ to map the video frames into embeddings $\rmE$ (in the space of the word embeddings) and Vicuna-7B-v1.0~\citep{vicuna2023} is used as $f_{llm}$.

\subsection{Frame-Scalable Encoding}
\label{sub:frame_scalable_encoding}
\begin{figure}
    \centering
    \includegraphics[width=\linewidth]{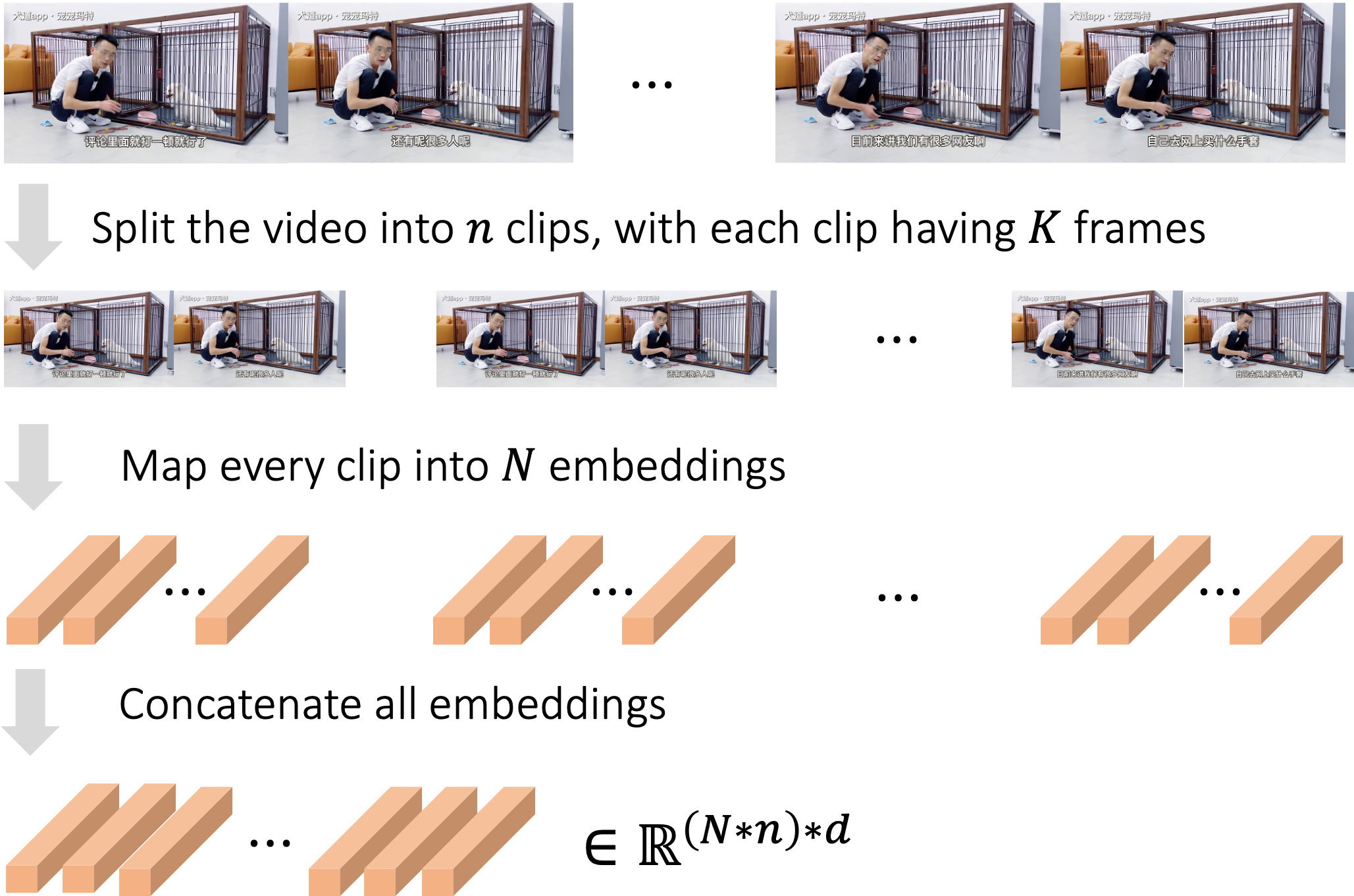}
    \caption{Illustration of Frame-Scalable Encoding. The process begins by segmenting the video into several clips. Subsequently, each clip is transformed into a set of $\cliptokens$ embeddings. These embeddings are then concatenated sequentially, forming a comprehensive input stream for the Large Language Model (LLM).}
    \vspace{-10pt}
    \label{fig:framework}
\end{figure}

 We observe a limitation in previous approaches that potential over-compression on given (long) videos can happen. To address this issue, we propose a novel encoding strategy, Frame-Scalable Encoding (FSE), based on the intuition that the number of embeddings allocated for video representation should be sufficient to cover the information within the video. 
The framework is shown in Figure \ref{fig:framework}. Specifically,
given a long video $\rmV$, 
FSE requires the video to be segmented into a series of clips $\rmV_1, \cdots, \rmV_{\numofclips}$, each bounded by a predefined maximum frame count. Then each clip is converted into a designated number of embeddings with Eq.(\ref{eq:video2embedding}): 
\begin{equation}
\label{eq:clipvideo2embedding}
    \rmE_1, \cdots, \rmE_\numofclips = f_{vid}(\rmV_1), \cdots, f_{vid}(\rmV_\numofclips)
\end{equation}
Here each embedding $\rmE_i \in \mathbb{R}^{\cliptokens \times d}, i\in \{1,\cdots,\numofclips\} $, where $d$ is the hidden dimension of the LLM.  
% Specifically, for a given video $\rmV$, we segment it into $\numofclips$ clips, and each clip is transformed into $\cliptokens$ embeddings.
Then we concatenate all embeddings $\rmE_1,\cdots,\rmE_n$ into the final representation $\rmE_{FSE} \in \mathbb{R}^{(n*N) \times d}$. 
% Consequently, the representation of $\rmV$ comprises $\numofclips * \cliptokens$ embeddings, 
Thus the representation $\rmE_{FSE}$ comprises $\numofclips * \cliptokens$ embeddings. When the video gets longer, we could obtain more clips ($\numofclips$ would increase), leading to more embeddings and
effectively mitigating the risk of over-compression. 
To determine how many clips we need, we propose the following equation: 
\begin{equation}\label{eq:numofclips}
\numofclips = \lceil \duration / \clipframes \rceil
\end{equation}
where $\duration$ denotes the video's duration (measured in seconds), ensuring a minimum of one frame per second is utilized. 
As the backbone model VideoChat2 is trained with the embeddings $\rmE \in \mathbb{R}^{N \times d}$ (i.e., only $\cliptokens$ embeddings), it may struggle with our embeddings $\rmE_{FSE}$ 
which comprises $\numofclips * \cliptokens$ embeddings. 
Thus we fine-tune the backbone model with the FSE embeddings. During training, due to the limitation of the resources and the constraint of maximal positional embeddings, we specify a maximum number of clips $\maxnumofclips$ and only sample $\maxnumofclips$ clips when videos get long, resulting in $\maxnumofclips * \clipframes$ frames. 
During inference, for longer videos, we can keep the strategy from training, i.e., only sample $\maxnumofclips$ clips. However, we propose a more optimal solution and explain mode details in the subsequent section.

% Under the circumstance where $\lceil \duration / \clipframes \rceil$ exceeds  $\maxnumofclips$, during training, we only sample $\maxnumofclips$ clips, equating to $\maxnumofclips * \clipframes$ frames. During inference, we propose a more refined solution discussed in the subsequent section. To enable the model to understand multiple video clips, we need to fine-tune the model on the video instruction dataset. 

% one temporary solution is to sample only $\maxnumofclips$ clips, equating to $\maxnumofclips * \clipframes$ frames, while a more refined solution is discussed in the subsequent solution. 

\begin{figure*}
    \centering
    \includegraphics[width=1.0\linewidth]{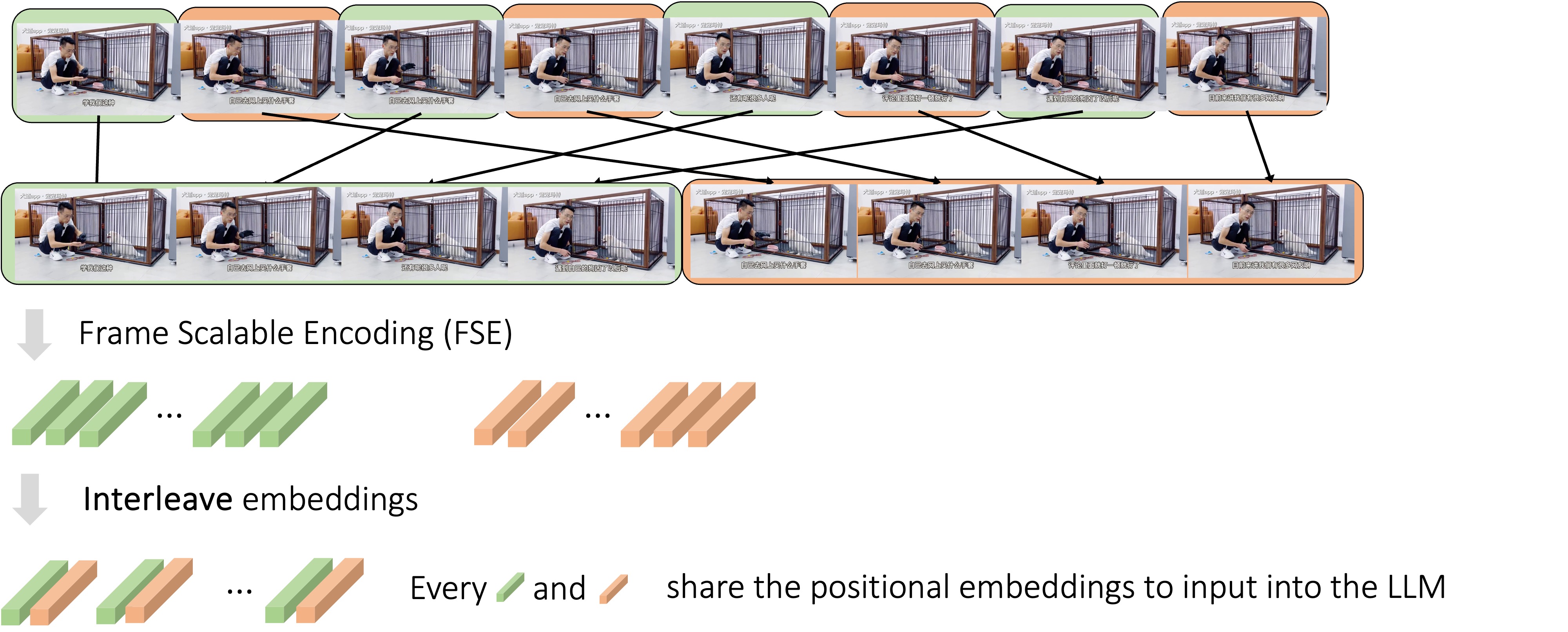}
    \caption{Illustration of Interleaved Frame Encoding (IFE). We show the example with interleaving factor $\numinter$ being two. We first split the whole video into $\numinter$ groups. Then we convert each part into embeddings separately. With all the embeddings, we interleave them with every $\numinter$ embeddings sharing the same positional embedding.}
    \label{fig:interleaving_framework}
    % \vspace{-8pt}
\end{figure*}

\subsection{Interleaved Frame Encoding}
\label{sub:interleaved_frame_encoding}
Although Frame-Scale Encoding (FSE) could mitigate the over-compression to some extent, it may introduce another challenge: when FSE is applied to excessively long videos, we may obtain an unwieldy number of embeddings from Eq.(\ref{eq:clipvideo2embedding}). %This is due to the linear correlation between video length and the number of generated embeddings. 
When there are overly many embeddings, it may surpass the maximum positional embeddings of the LLM. It may also encounter a problem that the embeddings during the inference is longer than the embeddings seen during training, leading to out-of-distrubition (OOD) problems. 
As discussed in \S~\ref{sub:frame_scalable_encoding}, a suboptimal solution could be limiting the clip numbers to be less than $\maxnumofclips$, but this approach may still suffer from over-compression identified in \S~\ref{introduction} 
%as the number of embeddings could be limited by $\maxnumofclips$,
as the number of embeddings is not scalable w.r.t. the video length, thereby limiting the effectiveness of FSE.

To tackle this challenge, we propose \textbf{Interleaved Frame Encoding (IFE)}. 
% Similar to Self-Extending LLM~\citep{self-extend-LLM}, 
IFE employs a repetition factor, $\numinter$ for the positional embeddings. Therefore, the positional embeddings are repeated at a predefined interval, $\numinter$, so that the sampled embeddings are within the range of training length,  mitigating the OOD issues or potential risk of surpassing maximum positional embeddings of the LLM.  
The process of IFE is depicted in Figure \ref{fig:interleaving_framework}. As shown in the figure, we split the video into $\numinter$ groups $\rmV_1, \cdots, \rmV_{\numinter}$. Each group is converted into embeddings $\rmE_{FSE,1}, \cdots, \rmE_{FSE,\numinter}$ using FSE techniques. These embeddings are fed into the LLM with the same positional embeddings applied to each group.
% We split the video into several groups and the corresponding positions in each segments share positional embeddings. 
One property we wish to include is that even when one group of the video is processed in isolation, without interleaving, IFE should align with using FSE with $\maxnumofclips$ clips.   
To achieve this, 
the video is divided into $\numinter$ groups in an interleaved way (shown in the Figure \ref{fig:interleaving_framework}). Then each group is encoded into embeddings independently. After this encoding phase, all embeddings are interleaved before being fed into the LLM. As illustrated, maintaining only one group (e.g., removing the right part in Figure \ref{fig:interleaving_framework}) effectively simulates the FSE scenario, sampling only the frames in one group (green frames in the example). Incorporating additional groups is intuitively expected to enhance the understanding of the video. 

For IFE, we determine the interleaving factor $\numinter$ by the following equation: 
\begin{equation}\label{eq:interleaving_factor}
    \numinter = \lceil \lceil \duration / \clipframes \rceil / \maxnumofclips \rceil.
\end{equation}
The intuition behind Eq.(\ref{eq:interleaving_factor}) is to make sure the number of clips in each group is less than $\maxnumofclips$ while maintaining the total amount of frames sampled could cover the whole video. 
Then we could sample $\sampledframes$ frames from the given video:
\begin{equation}
    \sampledframes = \lceil \lceil \duration/\clipframes \rceil / \numinter\rceil * \numinter * \clipframes
\end{equation}
Thus the number of the clips would be:
\begin{equation}
    \numofclipsinter = \lceil \lceil \duration/\clipframes \rceil / \numinter\rceil * \numinter
\end{equation}
With this strategy, we could sample $\sampledframes$ frames from the video which could cover the whole video,  as a result to have more than one frame per second, ensuring effective representation of the video.

\section{Experiments}

\subsection{Implementation Details}
We initialize our model from VideoChat2~\citep{li_videochat}. We set the learning rate as 2e-6, with warmup epochs=$0.3$, num\_epochs=$1$, scheduler=$cos$, optimizer=AdamW. The fine-tuning is performed on 4 NVIDIA-RTX-A6000 GPUs. For FSE, we finetune our model on the instruction dataset collected for training VideoChat2~\citep{li_videochat} with the detailed datasets shown in Appendix \S\ref{datasets_details}. 

For \ours, we use Eq.(\ref{eq:numofclips}) to determine the number of frames to sample, and  encode every $\clipframes$ frames into $\cliptokens$ embeddings, where $\clipframes=16, \cliptokens=96$. During the training, we specify $\maxnumofclips = 10$. Thus if the video length $\duration$ is shorter than $\maxnumofclips * \clipframes = 160$, we do not need IFE and only FSE is turned on, whereas if the video length $\duration$ is longer than $160$, we determine the interleaving factor $\numinter$ with Eq.(\ref{eq:interleaving_factor}) and then perform the IFE process. 

\subsection{Experimental Setups}
We compare \ours with the following models: \\
\textbf{VideoChat2}~\citep{li_videochat}: The backbone of our model without FSE and IFE. We follow the implementation\footnote{\url{https://github.com/OpenGVLab/Ask-Anything/blob/main/video_chat2/mvbench.ipynb}} in VideoChat2 and sample $16$ frames from the given video regardless of the video length.\\
\textbf{Video-Llama}~\citep{videollama}: We exclude the audio modality here for fair comparison. Following the setting from the original implementation\footnote{\url{https://github.com/DAMO-NLP-SG/Video-LLaMA}} model, we use the Video-LLaMA-2-7B-Finetuned checkpoint and sample $16$ frames from each video.\\
\textbf{Video-ChatGPT}~\citep{video-chatgpt}: We use the same setup as in the official demo\footnote{\url{https://github.com/mbzuai-oryx/Video-ChatGPT/blob/main/docs/offline_demo.md}} and samples 100 frames from each video.

For the benchmarks, we adopt MVBench ~\citep{mvbench} and extend the videos with 
the Street-Scene \cite{street-scene} dataset to specific lengths (see \S\ref{sub:dataset_extension}). We select the following 4 out of 20 test sets from MVBench: Action Sequence(AS), Action Prediction(AP), Unexpected Action(UA), Object Interaction(OI), following the criteria detailed in  Appendix\S~\ref{sub:datasets_selection_criteria}. The average length of the videos from these subsets is 25.5 seconds. We extend the original videos to 100s, 300s, and 600s respectively.
All models are evaluated under the same protocol proposed in MVBench. The prompts used are summarized in Appendix \S\ref{sub:prompt}.
We also report the performance on all subsets of MVBench in Appendix \S\ref{sub:all_subsets}.

% All models are evaluated under the protocol in MVBench\cite{mvbench}. For our own models, we ensemble 16 frames in one video token, which corresponds to 96 text tokens. 
% By joining $n$ video tokens, $16*n$ frames or $96*n$ text tokens will be used. For better alignment on frame-rate that the model see, we use $\lceil L / 16 \rceil * 16$ frames for video of L seconds if the total frame number doesn't exceed the maximal number of frames the model has seen in training stage($N_{max}$); when the length of video (or the projected number of frames) exceeds $N_{max}$, we compromise the sample rate and choose a frame number of $N_{max}$.

\subsection{Overall Performance Comparison}
\label{sub:overall_performance_comparison}
\begin{figure}[ht]
\centering
    \includegraphics[width=0.8\linewidth]{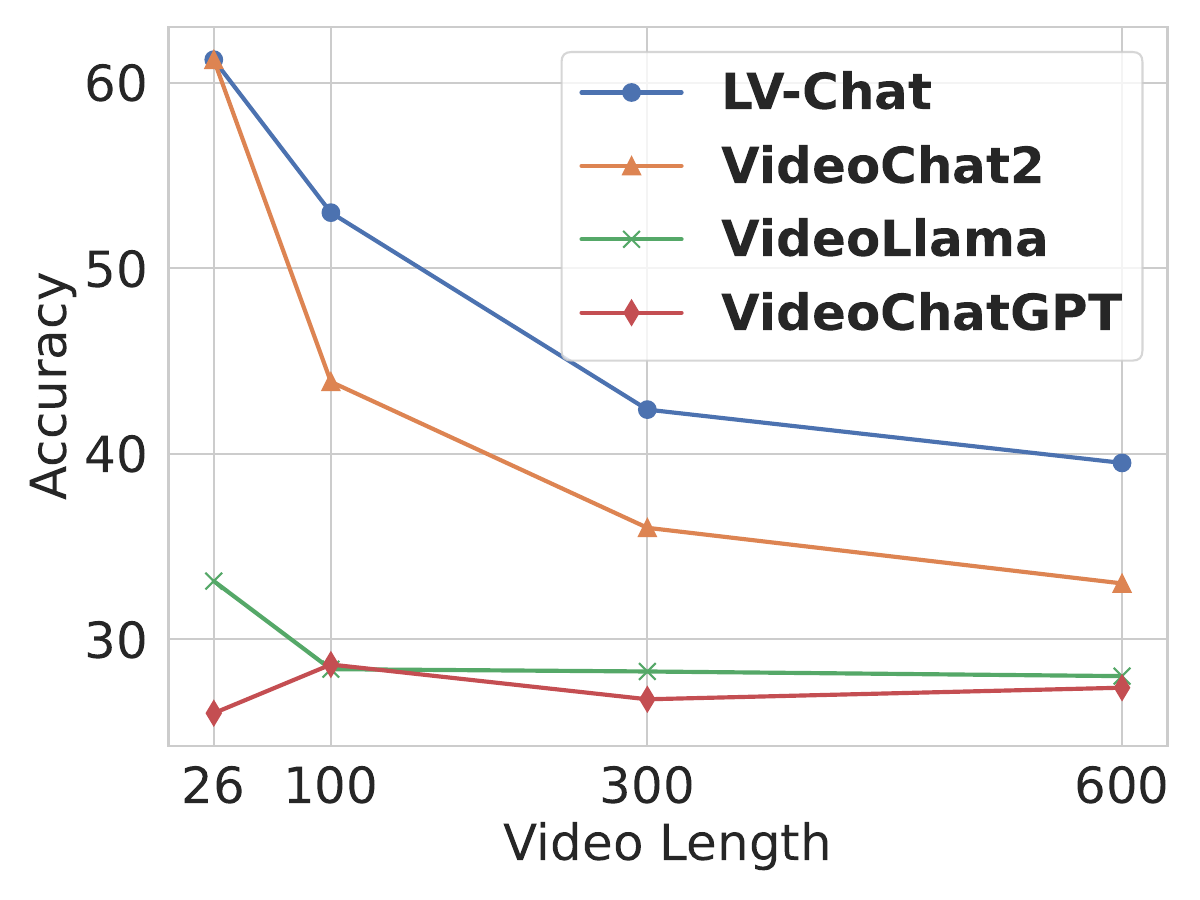}
    \caption{Average accuracies w.r.t different video lengths. ``26" is the average duration of videos across four datasets. 
    The IFE technique is not applied when videos are of lengths 26 and 100.}
    \label{fig:acc_wrt_video_length}
\end{figure}

We report the overall performance comparison in Table~\ref{tab:main}. From this table, we can observe that \ours outperforms the previous methods significantly on almost all datasets and in almost all settings. The results demonstrate that \ours could better extract the important information from the video even when the video becomes as long as 600s. We also summarize the average results of all datasets w.r.t. different video length in Figure \ref{fig:acc_wrt_video_length}. As shown in the figure, our model can achieve a compatible performance in terms of short videos but significant improvements on long videos.

%  The raw results are shown in Table ~\ref{tab:main}. We also shows the averaged results in Figure ~\ref{fig:acc_wrt_video_length}
% .
% We could observe that: (1) \ours could achieve the best performance in almost all datasets in all three settings; (2) With Interleaved Frame Encoding (IFE), \ours could achieve even higher performance than without IFE. 

% Table generated by Excel2LaTeX from sheet 'main result'
\begin{table*}[htbp]
  \centering
    % \resizebox{\textwidth}{!}{
    \begin{tabular}{l|rrrr|rrrr|rrrr}
    \toprule
& \multicolumn{4}{c|}{100s} & \multicolumn{4}{c|}{300s} & \multicolumn{4}{c}{600s} \\
%\midrule
& AS & AP & UA & OI & AS & AP & UA & OI & AS & AP & UA & OI \\
    \midrule
VideoChatGPT & 30    & 23    & 34    & 27.5  & 27.5  & 25.5  & 28    & 26    & 26    & 27    & 30    & 26.5 \\
VideoLlama & 24    & 23.5  & 39    & 27    & 25.5  & 23.5  & 38    & 26    & 23.5  & 25    & 37.5  & 26 \\
% videochat2(16*1) & 38.5  & 33    & 46.5  & 57.5  & 30.5  & 29    & \textbf{45} & 39.5  & 28.5  & 23    & \textbf{41.5} & 39 \\
VideoChat2 & 38.5  & 33    & 46.5  & 57.5  & 30.5  & 29    & \textbf{45} & 39.5  & 28.5  & 23    & \textbf{41.5} & 39 \\ \midrule
% videochat2(16*10) & 35.5  & 33.5  & 36.5  & 43    & 32    & 28.5  & 28.5  & 39    & 27    & 28    & 28    & 35.5 \\
% \ours(16*10) & \textbf{53.5} & \textbf{45.5} & \textbf{47} & \textbf{66} & 41    & \textbf{38.5} & 38.5  & 47    & 34.5  & 30.5  & 38.5  & 46 \\
% \ours+IFE(16*10) & -   & -   & -   & -   & \textbf{42.5} & 37.5  & 37    & \textbf{52.5} & \textbf{37} & \textbf{34} & 38.5  & \textbf{48.5} \\
\ours & \textbf{53.5} & \textbf{45.5} & \textbf{47} & \textbf{66} & \textbf{42.5} & \textbf{37.5}  & 37    & \textbf{52.5} & \textbf{37} & \textbf{34} & 38.5  & \textbf{48.5} \\
\bottomrule
    \end{tabular}%
    % }
  \caption{Results on QA datasets extended from MVBench. 
  % For videos of 100 seconds, $16*7$ tokens are used in videochat2(16*10) and our models.
  The interleaving factor $\numinter$ is set to be $2$ for videos of length 5 min and $4$ for videos of length 10 min. All models are evaluated using MVBench's protocol. }
  \label{tab:main}%
\end{table*}%

\begin{table*}[htbp]
  \centering
    % \resizebox{\textwidth}{!}{
    \begin{tabular}{l|rrrr|rrrr|rrrr}
    \toprule
& \multicolumn{4}{c|}{100s} & \multicolumn{4}{c|}{300s} & \multicolumn{4}{c}{600s} \\
%\midrule
& AS & AP & UA & OI & AS & AP & UA & OI & AS & AP & UA & OI \\
    \midrule
% videochat2(16*10) & 35.5  & 33.5  & 36.5  & 43    & 32    & 28.5  & 28.5  & 39    & 27    & 28    & 28    & 35.5 \\
% \ours(16*10) & \textbf{53.5} & \textbf{45.5} & \textbf{47} & \textbf{66} & 41    & \textbf{38.5} & 38.5  & 47    & 34.5  & 30.5  & 38.5  & 46 \\
% \ours+IFE(16*10) & -   & -   & -   & -   & \textbf{42.5} & 37.5  & 37    & \textbf{52.5} & \textbf{37} & \textbf{34} & 38.5  & \textbf{48.5} \\
\ours & \textbf{53.5} & \textbf{45.5} & \textbf{47} & \textbf{66} & \textbf{42.5} & 37.5  & 37    & \textbf{52.5} & \textbf{37} & \textbf{34} & \textbf{38.5}  & \textbf{48.5} \\
w/o IFE & - & - & - & - & 41    & \textbf{38.5} & \textbf{38.5}  & 47    & 34.5  & 30.5  & 38.5  & 46 \\
w/o IFE, w/o FSE & 35.5  & 33.5  & 36.5  & 43    & 32    & 28.5  & 28.5  & 39    & 27    & 28    & 28    & 35.5 \\
\bottomrule
    \end{tabular}%
    % }
  \caption{Ablation Study. We exclude IFE and FSE from \ours to study the effectiveness of these techniques. }
  \label{tab:ablation_study}%
\end{table*}%

% Appendix: \\
% Augment StreetScene; \\
% Augment Animal Motion / Human face; \\

% \begin{table*}[h!]
%     \centering
%     \begin{tabular}{ccccccc}
%     \toprule
%          & Base Video ($\sim 30$s) & Aug ($\sim 100$s) & Aug ($\sim 300$s) & Aug ($\sim 600$s) \\
%     \midrule
%         VideoChatGPT & \\
%         Video-Llama & \\
%         VideoChat2 & 71.5 & 52.5 & 36.0 & 34.0 \\
%         \ours & & & 49.5 & \\
%     \bottomrule
%     \end{tabular}
%     \caption{Experiments on Object Interaction. All settings are 10 clips with base frame number 8.}
%     \label{tab:exp_with_oi}
% \end{table*}

% \subsection{TVQA}

% \subsection{Long Video Chapter (Low Priority)}
% (VidChapter)

\subsection{Ablation Study of \ours}
We aim to study the effects of the finetuning of FSE (\S~\ref{sub:frame_scalable_encoding}) and the IFE technique (\S~\ref{sub:interleaved_frame_encoding}). Thus we exclude these two parts in \ours to check the performance on the benchmarks. The results are reported in Table \ref{tab:ablation_study}. From the table, we can observe that without IFE and FSE, the performance of \ours dropped, demonstrating the necessity of both FSE and IFE in terms of long video understanding.

\subsection{Model Analysis of \ours}
As the performance of \ours across the Action Sequence (AS) and Object Interaction (OI) datasets are most pronounced, we focus on these two datasets to study the efficacy and properties of \ours. 

\subsubsection{\ours can handle more embeddings}
\label{ssub:lv_chat_can_handle_more_embeddings}

\begin{figure}[ht]
    \centering
    \vspace{-10pt}
    \subfigure[Action Sequence]{\label{fig:action_sequence_acc_wrt_num_tokens}\includegraphics[width=0.49 \linewidth]{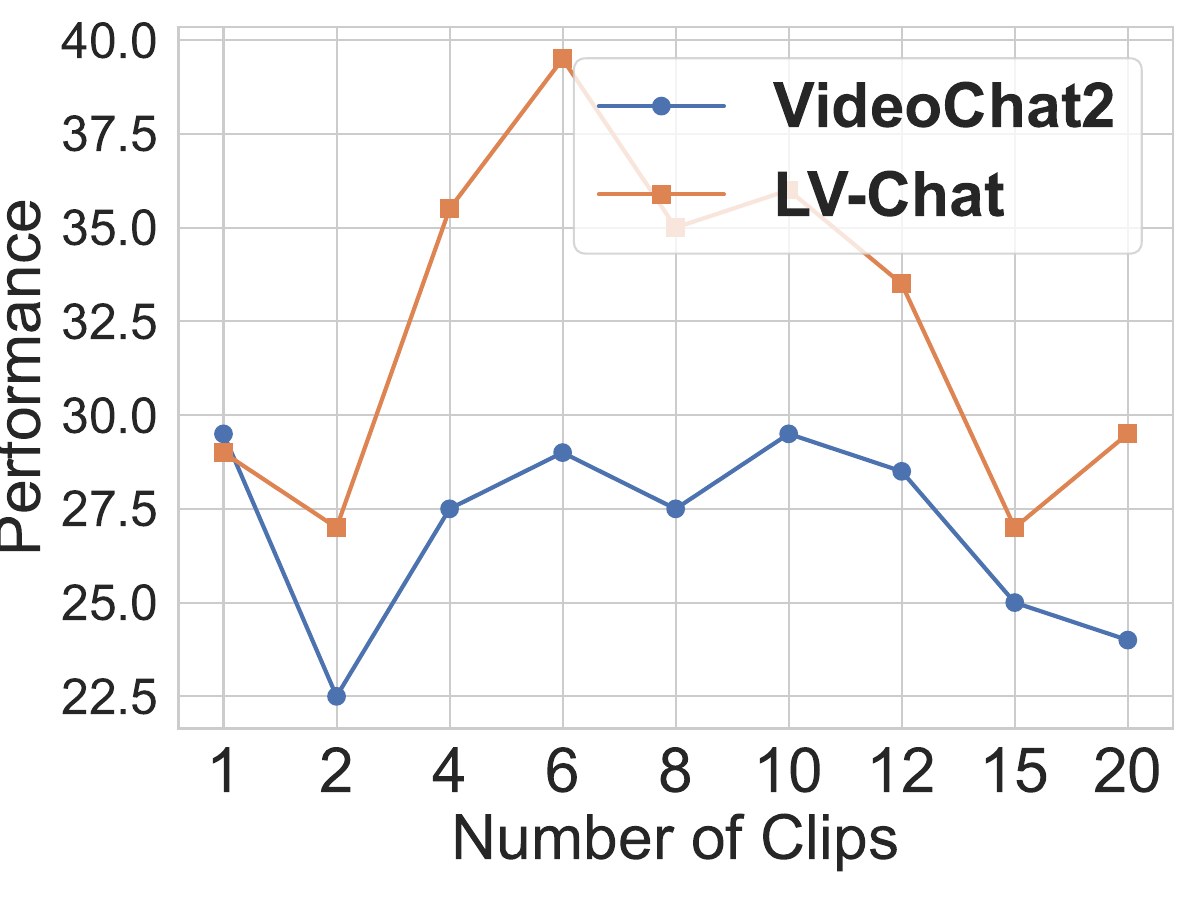}}
    \subfigure[Object Interaction]{\label{fig:object_interaction_acc_wrt_num_tokens}\includegraphics[width=0.49 \linewidth]{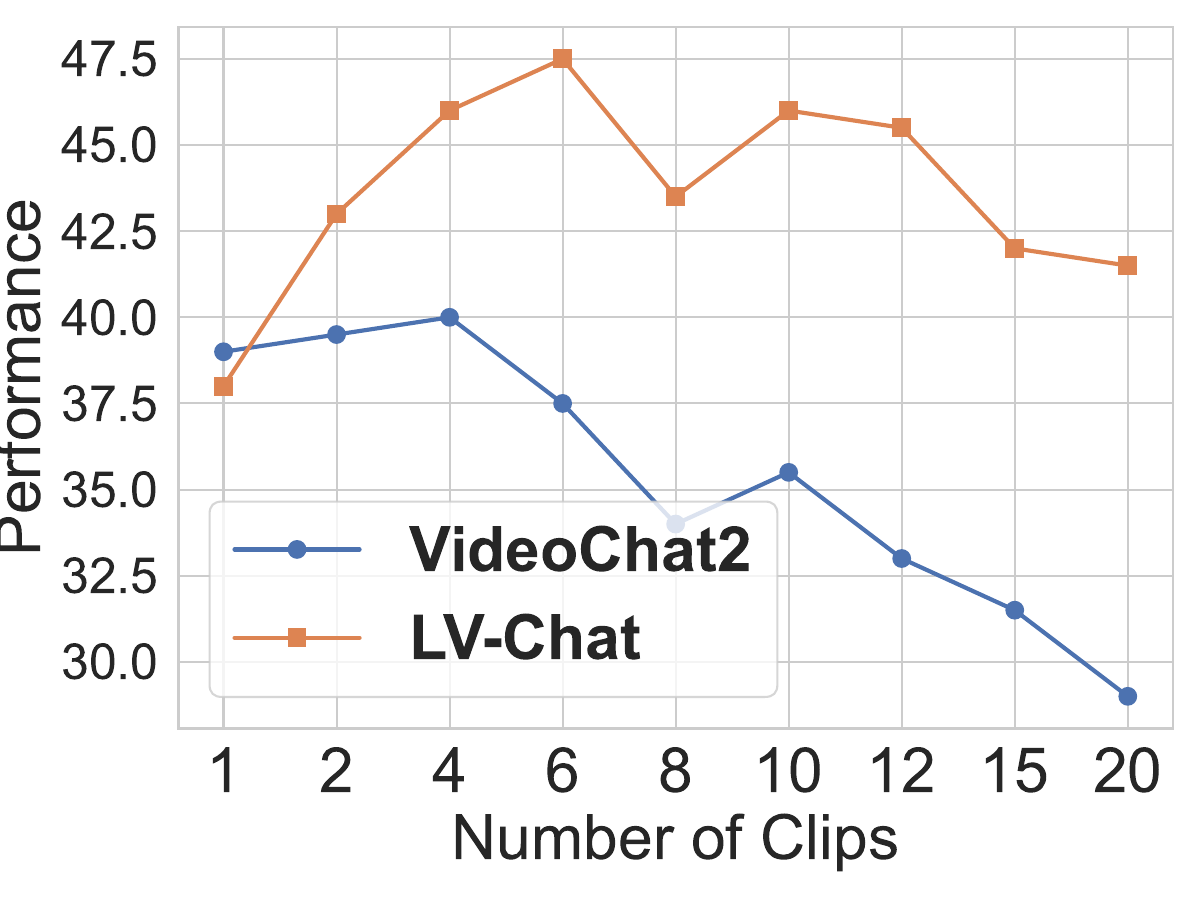}}
    \caption{Accuracies w.r.t. the number of tokens}
    \label{fig:more_tokens_helps}
    \vspace{-10pt}
\end{figure}

Our investigation aims to evaluate the performance difference between \ours and our backbone about the number of clips, $\numofclips$. For this purpose, experiments were conducted with $\numofclips$ ranging from 1 to 20, under a consistent video duration of 600 seconds.
%following the dataset conditions detailed in \S\ref{sub:overall_performance_comparison}. 
For instance, when $\numofclips=4$, we sample $\clipframes*4$ frames out of the entire video. 
The findings, illustrated in Figure~\ref{fig:more_tokens_helps}, reveal two key insights: (1) \ours consistently surpasses the baseline performance across all tested clip counts, which demonstrates the robustness and enhanced capacity for longer video understanding; (2) We noticed that as we increase the number of clips, \ours's performance gets better up to a certain point. Specifically, the model performs best with 6 clips. If we keep adding more clips beyond this number, up to 12, the performance barely drops. However, once we go over 12 clips, the performance begins to drop. This trend suggests that having too many clips doesn't always help as the model is trained with limited number of clips. This finding supports our earlier discussion about the challenges of matching the model's training experience with its usage in real-world scenarios, where the data it encounters can vary widely.

% The results show that our fine-tuned model consistently outperforms the baseline when it's given 2-12 tokens, while the baseline extending the token number alone doesn't help the baseline perform better.
% Besides, in both Figure ~\ref{fig:action_sequence_acc_wrt_num_tokens} and Figure ~\ref{fig:object_interaction_acc_wrt_num_tokens}, there is a similar tendency that the accuracy increases as the number of tokens increases and starts to drop when the number of tokens reaches a certain point. Interestingly, our fine-tuned model performs best on both datasets with 6 video tokens whilst the model is trained with 10 video tokens.

\subsubsection{Effectiveness of IFE}
\begin{figure}[ht]
    \centering
    %\subfigure[]{
        
        \includegraphics[width=0.8\linewidth]{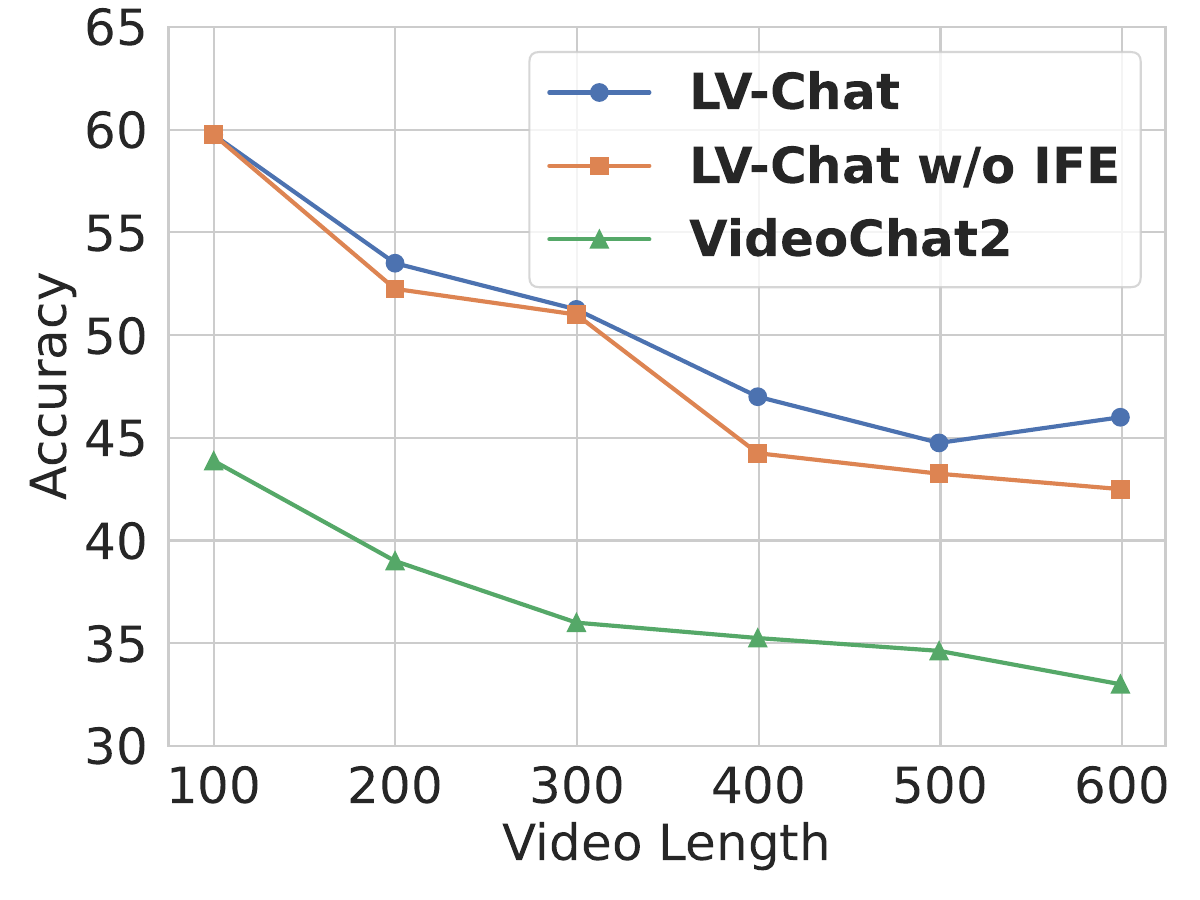}
    %}
    % \subfigure[Accuracy w.r.t different interleaving times]{
    % \label{fig:acc_wrt_interleaved_times}
    % \includegraphics[width=0.45\linewidth]{figures/acc_vs_interleave_times(16x10).pdf}
    % }
    \caption{IFE effectiveness on QA datasets.}
    \label{fig:effectiveness_of_ife}
    \vspace{-1.em}
\end{figure}

To assess how well Interleaved Frame Encoding (IFE) works, we tested it on videos of varying lengths: 100s, 200s, 300s, 400s, 500s, and 600s. For each video length, we adjusted the interleaving factor $\numinter$ from 1 to 6, respectively. This setup aligns with our previous finding that \ours shows optimal performance with up to 6 clips (as detailed in \S~\ref{ssub:lv_chat_can_handle_more_embeddings}). The results, summarized in Figure \ref{fig:effectiveness_of_ife}, indicate a clear trend: incorporating IFE improves the model's performance. Notably, as the video length increases, the benefit of using IFE becomes even more pronounced. Detailed performance metrics across four datasets are provided in \S~\ref{sub:detailed_results_for_the_effectiveness_of_ife}.

\begin{table*}[t]
    \centering
    \resizebox{\linewidth}{!}{
    \begin{tabular}{l|cccc|cccc|cccc}
    \toprule
    & \multicolumn{4}{c|}{100s} & \multicolumn{4}{c|}{300s} & \multicolumn{4}{c}{600s} \\
    & AS & AP & UA & OI & AS & AP & UA & OI & AS & AP & UA & OI \\
    \midrule
     % \ours & 48.5 & 44.0 & 42.5 & 61.0 & 42.5 & 35.5 & 36 & 49.5 & 34 & 32 & 34.5 & 49 \\
      % \ours(8*10) $+$ IFE & - & - & - & - & \textbf{43.5} & 37 & 33.5 & 50 & 34 & 32 & 34.5 & 49 \\
      \ours ($\clipframes=8$) & 48.5 & 44.0 & 42.5 & 61.0 & \textbf{43.5} & 37 & 33.5 & 50 & 34 & 32 & 34.5 & 49 \\
      \ours w/o IFE ($\clipframes=8$) & - & - & - & - & 42.5 & 35.5 & 36 & 49.5 & 34 & 32 & 34.5 & 49 \\
      \ours ($\clipframes=16$) & \textbf{53.5} & \textbf{45.5} & \textbf{47} & \textbf{66} & 42.5 & 37.5  & 37    & \textbf{52.5} & \textbf{37} & \textbf{34} & \textbf{38.5}  & \textbf{48.5} \\ 
      \ours w/o IFE ($\clipframes=16$) & - & - & - & - & 41    & \textbf{38.5} & \textbf{38.5}  & 47    & 34.5  & 30.5  & 38.5  & 46 \\
 % \ours(16*10) & \textbf{53.5} & \textbf{45.5} & \textbf{47} & \textbf{66} & 41    & \textbf{38.5} & \textbf{38.5}  & 47    & 34.5  & 30.5  & 38.5  & 46 \\
% \ours (16*10) $+$ IFE & -   & -   & -   & -   & 42.5 & 37.5  & 37    & \textbf{52.5} & \textbf{37} & \textbf{34} & \textbf{38.5}  & \textbf{48.5} \\
     \bottomrule
    \end{tabular}}
    \caption{Ablation study with different $\clipframes$ on long-video question answering benchmarks. \textbf{Bold}: best results.}
    \label{tab:varying_clip_frames}
\end{table*}

\begin{table*}[t]
\centering
\begin{tabular}{l|cccc|c}
\toprule
&  \multicolumn{4}{c|}{TACoS(287s)} & \multicolumn{1}{c}{EgoSchema(180s)} \\
& \multicolumn{1}{l}{Rouge1} & \multicolumn{1}{l}{Rouge2} & \multicolumn{1}{l}{RougeL} & \multicolumn{1}{l|}{RougeSum} & Accuracy \\
\midrule
 VideoLlama & 0.269 & 0.0490 & 0.196 & 0.193 & 0.284\\
VideoChatGPT & 0.263 & 0.0567 & 0.188 & 0.188 & 0.260\\
VideoChat2 & 0.261 & 0.0675 & 0.195 & 0.196 & 0.500 \\
\ours & 0.360 & 0.0920 & \textbf{0.244} & \textbf{0.246} & 0.554 \\
\midrule
\ours w/o IFE & \textbf{0.364} & \textbf{0.0931} & 0.244 & 0.245 & \textbf{0.560} \\
\bottomrule
\end{tabular}
    \caption{Evaluation on long-video caption generation datasets. \textbf{Bold}: best results.}
    \label{tab:real_world_data}
\end{table*}

\subsubsection{Varying $\clipframes$ in FSE}
As shown in our main experiments (\S~\ref{sub:overall_performance_comparison}), the number of frames per clip is set as $\clipframes=16$. We aim to show converting $16$ frames into $\cliptokens=96$ embeddings is not over-compression by checking the performance of \ours when we map every $\clipframes=8$ frames into $\cliptokens=96$ embeddings. The results with $\clipframes=8$ are reported in Table \ref{tab:varying_clip_frames}. From the table, we can see that $\clipframes=16$ performs better than $\clipframes=8$, showing that $\clipframes=16$ may have not led to over-compression as $\clipframes=8$ could not mitigate the potential over-compression problem. This observation is also partially observed in VideoChat2~\citep{li_videochat} where extracting 16 frames from the video and mapping them into 96 tokens generally perform better than extracting 8 frames.

\newcommand{\firstscene}{\textcolor{cyan}}
\newcommand{\secondscene}{\textcolor{orange}}
\newcommand{\scenewidth}{5cm}
\newcommand{\resultwidth}{9.5cm}
\newcommand{\figurewidth}{3.2cm}

\begin{table*}[ht]
\footnotesize
\centering
% \begin{tabular}{\textwidth}{c|c}
\begin{tabular}{c|c}
\toprule
Captioned scenes & Results \\
\midrule
\begin{tabular}{p{\scenewidth}}
\centering
\includegraphics[width=\figurewidth]{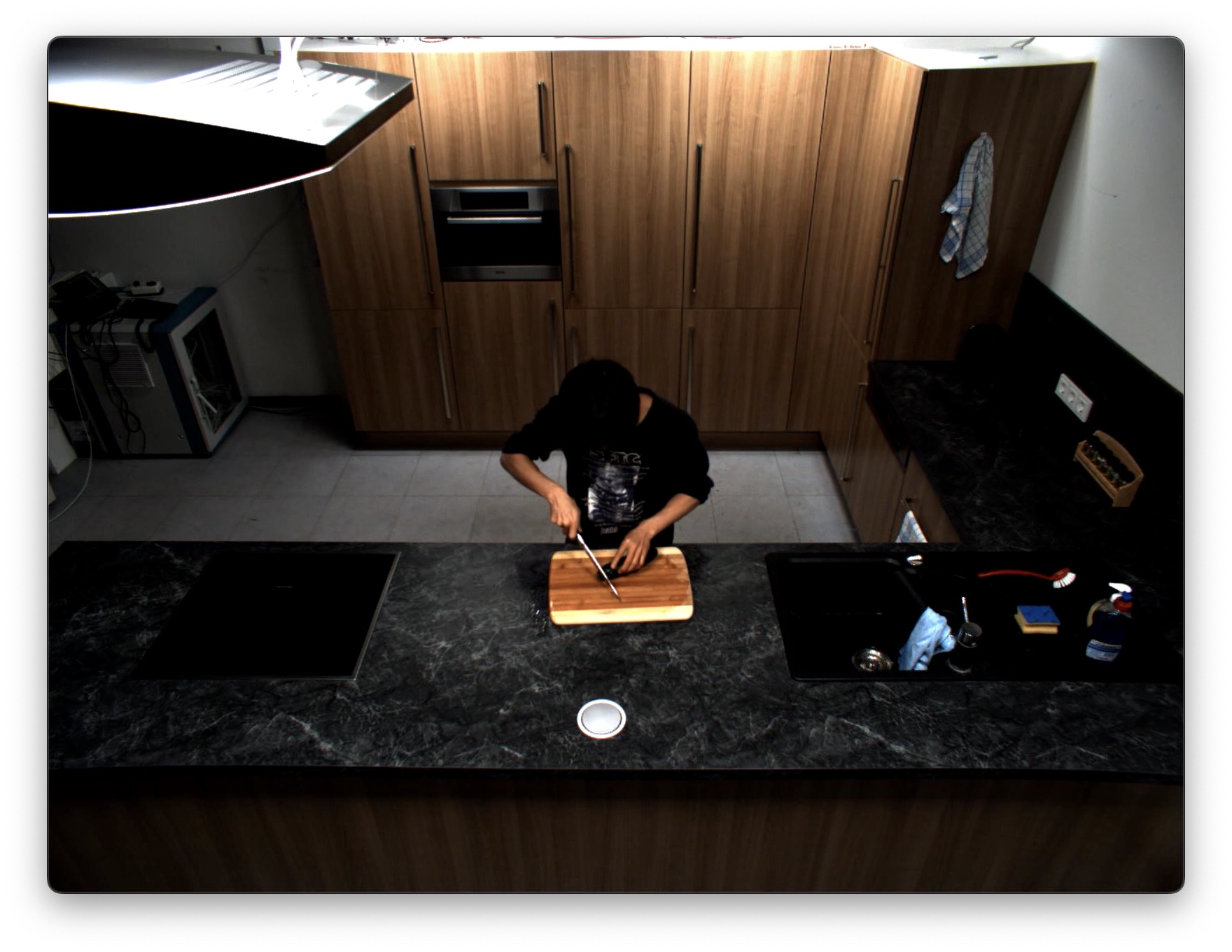} \\
\firstscene{He cut off ends of cucumbers.}\\
\includegraphics[width=\figurewidth]{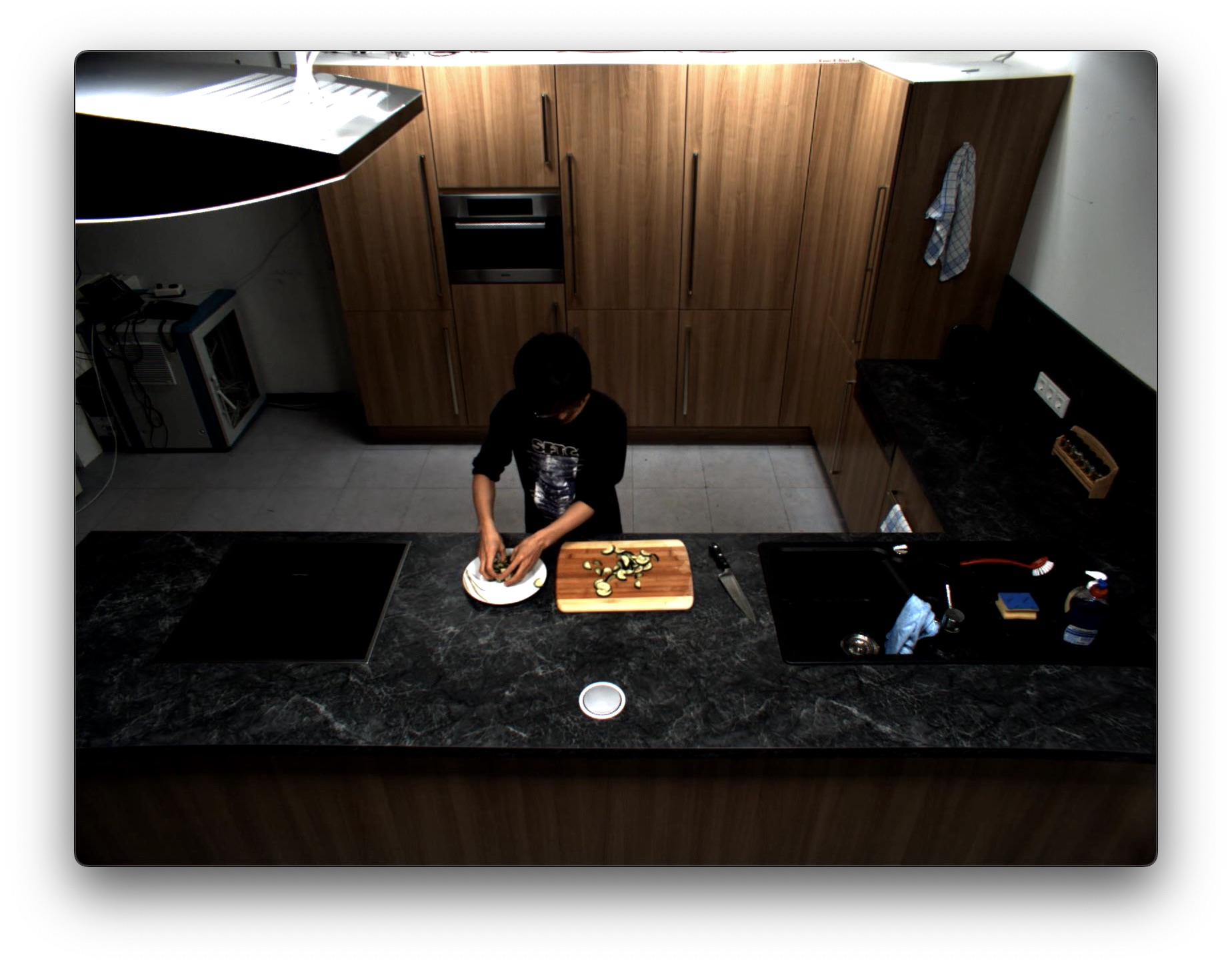}\\
\secondscene{He put cucumber on plate.}
\end{tabular}

 & \begin{tabular}{p{\resultwidth}}
\ours: ``The video shows a man standing in a kitchen and \firstscene{cutting a cucumber on a cutting board}. He then \secondscene{places the cucumber on a plate and walks away}. The kitchen has a counter with a sink and a refrigerator. The man is wearing a black shirt and is standing in front of a black and white fridge. The video is shot from an overhead angle, giving a bird's eye view of the kitchen...''\\
\midrule
     VideoChat2: ``The video shows a man \firstscene{cutting a cucumber on a cutting board}.''\\
     \midrule
     VideoChatGPT: ``The video shows a person \firstscene{cutting vegetables on a countertop} in a kitchen.''\\
     \midrule
VideoLlama: ``1. A man is standing in a kitchen with a black countertop and a wooden table. 2. He is wearing a black shirt and a black jacket. 3. He is preparing food in the kitchen. 4. He is using a knife to cut a piece of meat. 5. He is cutting the meat on a cutting board. 6. He is using a spoon to stir the food. 7. He is standing in the kitchen''
\end{tabular} \\
\midrule
% \begin{tabular}{c}
% s28-d51\\
% (11min11s)
% \end{tabular}
%\multirow{4}[0]{*}{s28-d51} % 11:11
 % & 
\begin{tabular}{p{\scenewidth}}
\centering
\includegraphics[width=\figurewidth]{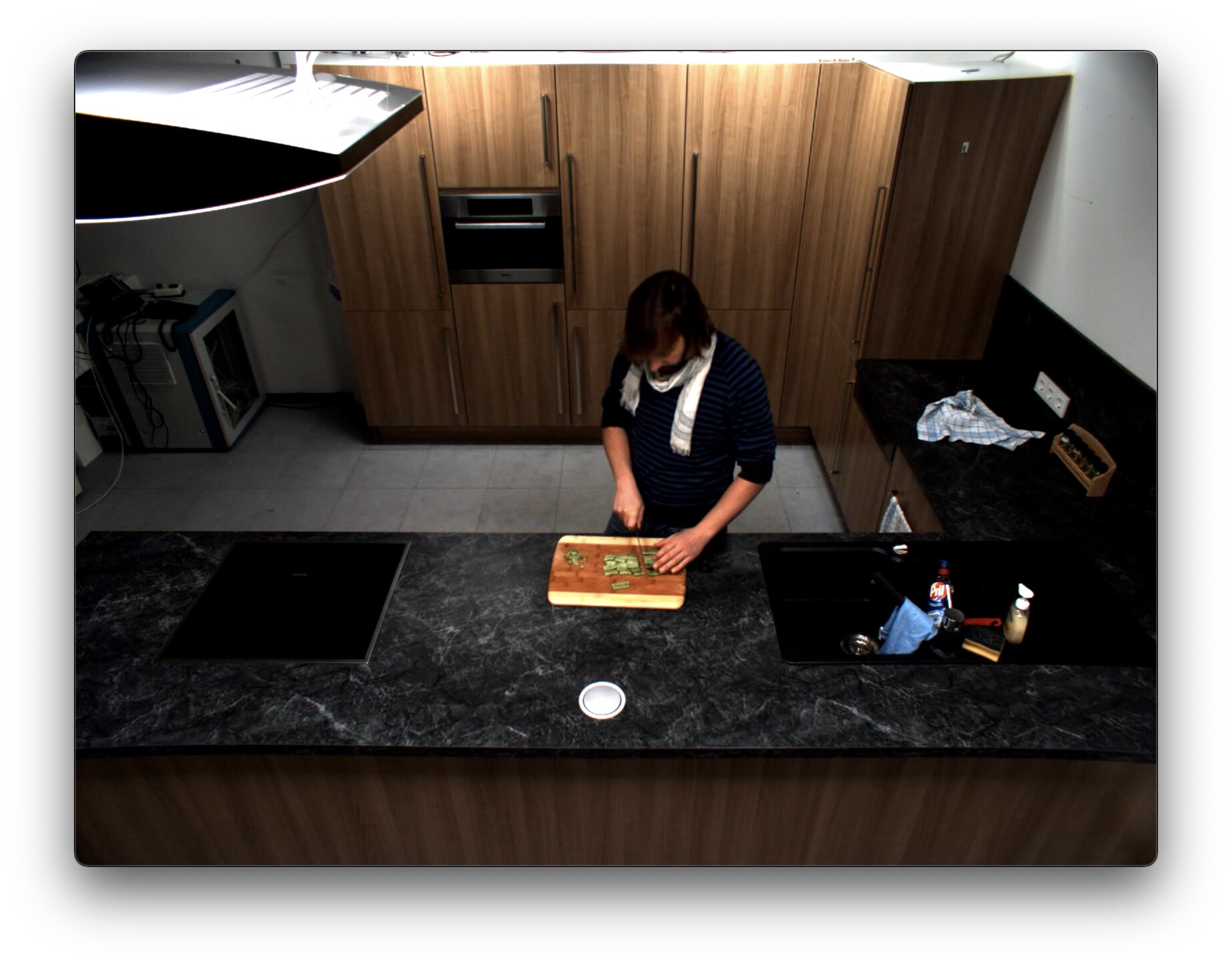} \\
\firstscene{The man slices the broad beans.}\\
\includegraphics[width=\figurewidth]{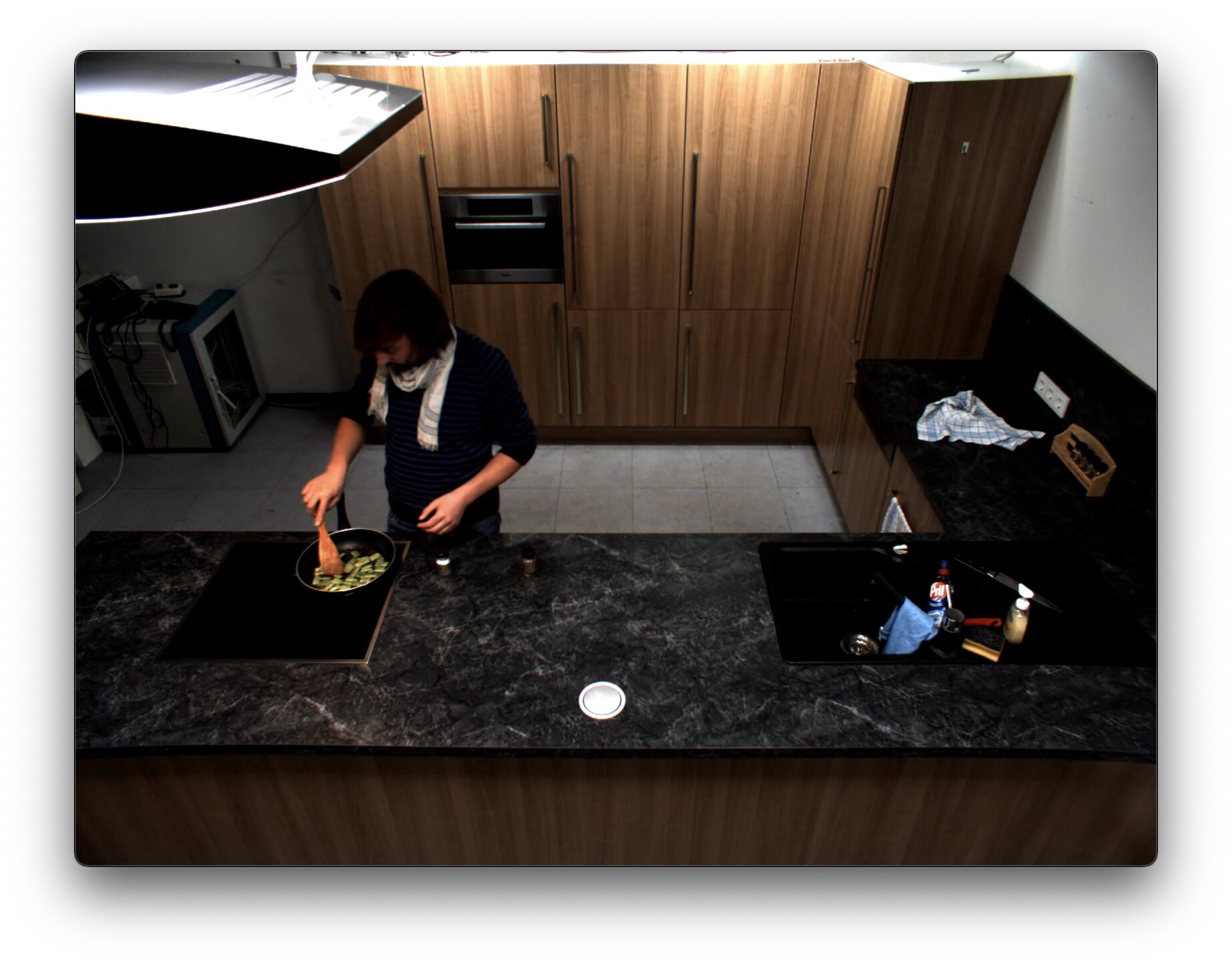} \\
\secondscene{The man continues to stir the beans as they cook.}
\end{tabular}
&
\begin{tabular}{p{\resultwidth}}
\ours: ``In the video, we see a man standing in a kitchen and preparing food. He is wearing a blue shirt and a white scarf around his neck. He is \firstscene{cutting vegetables on a cutting board and placing them in a pan}. He then \secondscene{adds some oil to the pan and starts cooking the vegetables}. The man is standing in front of a counter with a black and white marble top. There is a microwave oven on the counter, and a white bowl...'' \\
\midrule
VideoChat2: ``The video shows a man \firstscene{cutting vegetables on a cutting board}.'' \\
\midrule
VideoChatGPT: ``In the video, a woman is standing in a kitchen and preparing food. She is using a knife to \firstscene{chop vegetables on a cutting board}, and she is also using a spoon to stir something in a bowl. The video also shows her using a spoon to stir something in a bowl.''\\
\midrule
VideoLlama: ``The video shows a man and a woman in a kitchen. The man is standing in front of a stove while the woman is standing in front of a counter. They are both wearing blue shirts. The kitchen has a black countertop and a black stove. There is a brown wooden table in the kitchen. The man is holding a knife and the woman is holding a plate. They are both preparing food.'' 
\end{tabular}\\
\bottomrule
\end{tabular}
\caption{Two cases on the TACoS dataset of \ours compared with the baselines. The lengths of the two videos are 2 min 46 s and 11 min 11 s respectively. The highlighted parts are correct descriptions of actions.}
\label{tab:case_study}
\end{table*}

\subsection{Real World Datasets}
In this section, we evaluate the performances of the baselines and \ours on two real-world datasets (by real-world, we mean the videos are naturally long videos, instead of extending short videos with unrelated ones):\\
\textbf{TACoS}~\citep{tacos}: This is a dataset comprising 127 videos averaging 287s and human-annotated captions of critical timestamps in the video. We use OpenAI's GPT-4~\citep{openai2024chatgpt} to generate a reference summary from the labeled captions and conduct a human inspection (detailed in \S\ref{sub:prompt_tacos_summary}). Then all models are prompted to generate detailed descriptions and the ROUGE scores are calculated against the reference.\\
\textbf{EgoSchema}~\citep{egoschema}: Derived from Ego4D~\citep{ego4d}, it is a long-form video  question-answering dataset with an average length of 180; 500 samples with public released answers are used for evaluation. \\
For TACoS description generation, all models generate up to 100 new tokens using greedy search. For EgoSchema evaluation, we use the same settings as MVBench. The results are reported in Table \ref{tab:real_world_data}. From the table, we can observe: (1) \ours outperforms all the other baselines by a large margin. Note that we have not done any fine-tuning on the datasets being tested (2) Although IFE does not provide any improvement over EgoSchema, we think one potential reason is that the lengths of the videos from this dataset are not sufficiently long ($160$ frames are sampled when IFE is not used, which is close to the average duration $180$s.)

\subsection{Case Study}
%Show some results here, it may take around half the page to show one or two videos and the question-answering results. 

%With a given example, show the results of different models.

%give a figure/table of e.g., 4 x 2, with images and questions and different answers. 

We compare \ours against the baselines on the TACoS and show results in Table~\ref{tab:case_study}. For each video, we choose two representative scenes and match them with the captions from the TACoS. 
%Due to the fact that \ours is fine-tuned from VideoChat2, in most cases they have similar performance. 
We observe there are a number of cases where VideoChat2 can only summarize the whole video in one sentence without any further detail. And VideoChatGPT suffers from the same issue. While VideoLlama generates longer answers generally, it often has strong hallucinations on the details of the video and gives far-off descriptions. In contrast, our model captures much more details, including the actions of the subject and the environment where the video was shot. In the cases we show, we also highlight the correct action descriptions that these models generate. All three baselines fail to correctly capture the actions of the person from both two scenes while \ours succeeds in describing both. More comparisons between our model and VideoChat2 are shown in Appendix~\ref{sec:gen_cases}.

\section{Conclusion}
% \vspace{-5pt}
In this study, we introduced Long Video Chat (\ours), a novel approach aimed at enhancing the comprehension capabilities of large language models (LLMs) for long video content. \ours has two innovative encoding strategies: Frame-Scalable Encoding (FSE) and Interleaved Frame Encoding (IFE). These techniques address the fundamental challenge of over-compression in video representation, a notable limitation in existing multimodal LLM frameworks when processing videos, particularly those exceeding one minute in duration. %maybe five minutes
%Through meticulously designed experiments, 
We evaluated \ours's performance in long video comprehension tasks, utilizing both curated datasets and real-world benchmark. Our findings demonstrate that \ours consistently surpasses previous methods in these settings. 
% One limitation of \ours is the fine-tuning using IFE, which, contrary to expectations, did not yield any enhancements. This may be attributed to the insufficiency of long videos in the current video instruction dataset. Consequently, future work includes the development of datasets with longer videos to achieve better performances via fine-tuning. 

\clearpage

\section{Limitations}
% During case study, we find that our model still struggles in figuring out a concrete name for the objects in a video and in identifying fine-grained actions, which also happen to the base model (VideoChat2) that we finetuned upon. Our proposed methods, however, are orthogonal to the visual capability of the base model and may benefit future models.
% During the case study, we observed that our model encounters challenges in accurately assigning specific names to objects within a video and in identifying fine-grained actions. Notably, these difficulties are also evident in the base model (VideoChat2). 
% It is noteworthy that our proposed methods are independent of the visual capabilities of the base model, suggesting potential benefits for future models.
One limitation of \ours lies in the fine-tuning stage with the IFE enabled, which, contrary to expectations, did not yield any enhancements. This may be attributed to the insufficiency of long videos in the current video instruction dataset. Consequently, future work includes the development of datasets with longer videos to achieve better performances via fine-tuning. 
Another limitation is that \ours is based on VideoChat2 which uses Vicuna-7B-v1.0 as the LLM, which may be inferior than the most advanced LLMs such as Vicuna-7B-v1.5. Thus another future work is to train a larger model with more advanced LLMs, enhancing the understanding capabilities. 

\bibliography{anthology,custom}
\bibliographystyle{acl_natbib}

\clearpage 
\appendix
% \onecolumn
\section{Notations}
All the notations are provided in Table \ref{tab:notations}. 
\begin{table}[h!]
    \centering
    \begin{tabular}{ccc}
    \toprule
        Symbols & Meanings \\
    \midrule
        $\duration$ & duration \\
        $\totalframes$ & total number of frames \\
        $\clipframes$ & number of frames in one clip \\
        $\cliptokens$ & number of tokens per clip \\
        $\sampledframes$ & number of sampled frames \\
        $\numofclips$ & number of clips \\
        $\maxnumofclips$ & max number of clips \\
        $\numinter$ & number of interleaved times \\
        $\numofclipsinter$ & number of clips in interleaved setting \\
    \bottomrule
    \end{tabular}
    \caption{Notations}
    \label{tab:notations}
\end{table}

\section{Experiment Settings}

\subsection{Instruction Tuning Dataset Details}

\label{datasets_details}
To fine-tune our model with FSE, we adopt the dataset collected by VideoChat2~\citep{li_videochat}, where there is 1.9M video instruction data in total\footnote{\url{https://github.com/OpenGVLab/Ask-Anything/blob/main/video_chat2/DATA.md}}. However, due to that some datasets are not accessible, we use a subset of this dataset:
\begin{itemize}
    \item VideoChat~\citep{li_videochat}, collected from InternVid~\citep{internvid}. 
    \item VideoChatGPT~\citep{video-chatgpt}, the original caption data is converted into conversation data by \cite{li_videochat}. 
    \item NExTQA~\citep{NExT-QA}, a multi-choice question answering dataset. 
    \item CLEVRER~\citep{clevrer}, an action prediction, multi-choice question answering dataset. 
\end{itemize}

\subsection{Datasets Selection Criteria}
\label{sub:datasets_selection_criteria}
By manually looking at the examples, we compiled a few rules that a valid set of data should satisfy:
\begin{enumerate}
    \item The baseline's performance drops as the target length of the extended video increases.
    \item The baseline's performance should be better than random guesses.
    \item Questions in the subset should not be greatly affected by video from Street-Scene.
    \item Video should not be too short compared to our target length.
    \item The questions in the subset should be answerable by a visual-only model. (i.e., the answers should not be all in the subtitles or the captions, leading to unanswerable questions based on visual data only)
\end{enumerate}

By applying these rules, we select four datasets (Action Sequence, Action Prediction, Unexpected Action, Object Interaction) that are valid for testing long video-language models.
\subsection{Dataset Extension}
\label{sub:dataset_extension}

Despite the variety of videos that MVBench\cite{mvbench} has. The average length of the four selected datasets are merely 25.5s, which can barely benefit from the capability of long-video models. To make use of these videos, we extend them with a second video sampled from the Street-Scene dataset\cite{street-scene}. The Street-Scene dataset contains 91 videos with 15 frames per second, and we select the first 54000 frames from the dataset, totaling an 1 hour video from which we sample the second video. 

The extension process is as follows:
\begin{enumerate}
    \item Set a target length of video $T$ that the model should see.
    \item For a original video $\mathit{v}$ of length $\mathcal{L}(v) < T$, we applies a hash function $\mathcal{H}$ (see below) to the file name $N_v$ of the video $v$ to get a integer $t_0$ that is between 0 and 3600, which will be used as the starting time of the second video. The hash function in python is:
    \begin{lstlisting}{python}
   def hashstr(s: str) -> int:
    return sum(ord(c) * 31 ** (i % 3) 
        for i, c in enumerate(s)) 
\end{lstlisting}
    \item  Draw a second video from the Street-Scene dataset that starts at $t_0 = \mathcal{H}(N_v)$ and ends at $t_0 + (T-\mathcal{L}(v))$.
    \item Choose a time point $t_1= \mathcal{H}(N_v + \text{":insert"})$ in the second video where we will insert the original video.
    \item Insert the original video at $t_1$ of the second video and returns the extended video.
\end{enumerate}

\subsection{GPT-4 TACoS summarization}\label{sub:prompt_tacos_summary}
We use the following content to query the ``GPT-4'' API from OpenAI on Oct.9th, 2023. The context is composed of human-labelled captions and their starting times. The template we use for prompting GPT-4 is:
\begin{lstlisting}
You are an assistant answering questions based on video contexts. Your answer should be based on the given contexts, but you can also infer the actual video content from the tag information and your common sense. The timed description is a description for the video at the given second. When describing, please mainly refer to the timed description. Don't create a video plot out of nothing. 
Contexts for the video: \{context\}
Question: Could you please describe what is happening in the video?
\end{lstlisting}

\begin{table*}[]
    \centering
    \begin{tabular}{l|llllllllll}
\toprule
                 & AS   & AP   & AA   & FA   & UA   & OE   & OI   & OS   & MD   & AL   \\
                 \midrule
VideoChat2       & 66   & 47.5 & 83.5 & 49.5 & 60   & 58   & 71.5 & 42.5 & 23   & 23   \\
VideoChatGPT     & 23.5 & 26   & 62   & 22.5 & 26.5 & 54   & 28   & 40   & 23   & 20   \\
VideoLlama      & 27.5 & 25.5 & 51   & 29   & 39   & 48   & 40.5 & 38   & 22.5 & 22.5 \\
LV-Chat          & 62.5 & 47   & 79.5 & 44   & 61.5 & 56   & 74   & 40.5 & 23.5 & 27   \\
\midrule
                & ST   & AC   & MC   & MA   & SC   & FP   & CO   & EN   & ER   & CI   \\
                 \midrule
VideoChat2 & 88   & 39   & 42   & 58.5 & 44   & 49   & 36.5 & 35   & 40.5 & 65.5 \\
VideoChatGPT     & 31   & 30.5 & 25.5 & 48.5 & 29   & 39.5 & 33   & 29.5 & 26   & 35.5 \\
VideoLlama      & 43   & 34   & 22.5 & 45.5 & 32.5 & 32.5 & 40   & 30   & 21   & 37   \\
LV-Chat          & 82   & 47.5 & 39.5 & 69.5 & 47   & 48.5 & 40   & 34.5 & 38.5 & 60  \\
\bottomrule
    \end{tabular}
    \caption{Model Performance on the original MVBench. The results of VideoChat2, VideoChatGPT and VideoLlama are from the MVBench repository (\url{https://github.com/OpenGVLab/Ask-Anything/blob/main/video_chat2/MVBENCH.md}).}
    \label{tab:mvbench_original}
\end{table*}
\begin{table*}[ht]
\footnotesize
\centering
\begin{tabular}{l|llllllllll}
\toprule
\multicolumn{11}{c}{Length 100s}                                                                   \\
\midrule
                          & AS   & AP   & AA   & FA   & UA   & OE   & OI   & OS   & MD   & AL   \\
\midrule
VideoChat2(16*1)          & 38.5 & 33   & 64.5 & 34   & 46.5 & 53   & 57.5 & 31.5 & 23.5 & 29   \\
VideoChat2(16*10)         & 35.5 & 33.5 & 41.5 & 29.5 & 36.5 & 54.5 & 43   & 38   & 19.5 & 22   \\
VideoChat2(8*10)          & 36.5 & 33   & 43   & 28   & 34.5 & 54   & 41.5 & 38   & 18.5 & 23   \\
VideoChatGPT              & 30   & 23   & 54.5 & 24   & 34   & 53.5 & 27.5 & 41   & 24.5 & 26.5 \\
VideoLlama               & 24   & 23.5 & 42.5 & 27   & 39   & 52.5 & 27   & 33   & 23.5 & 21   \\
LV-Chat(8*10)             & 48.5 & 44   & 52.5 & 28.5 & 42.5 & 55   & 61   & 34   & 20.5 & 29   \\
LV-Chat(16*10)            & 53.5 & 45.5 & 59.5 & 30   & 47   & 53   & 66   & 36.5 & 20.5 & 28   \\
\midrule
                          & ST   & AC   & MC   & MA   & SC   & FP   & CO   & EN   & ER   & CI   \\
                          \midrule
VideoChat2(16*1)          & 72   & 43.5 & 30.5 & 57.5 & 54   & 29   & 40   & 31   & 39.5 & 43.5 \\
VideoChat2(16*10)         & 40   & 39.5 & 22.5 & 37.5 & 58.5 & 26.5 & 38   & 24.5 & 30.5 & 39.5 \\
VideoChat2(8*10)          & 40   & 38   & 22.5 & 37   & 57.5 & 27   & 41   & 25.5 & 32   & 44.5 \\
VideoChatGPT              & 40   & 30   & 29   & 36.5 & 48.5 & 21   & 36   & 28.5 & 29   & 39   \\
VideoLlama               & 32.5 & 29   & 28   & 41.5 & 45.5 & 29   & 34.5 & 30   & 25   & 35.5 \\
LV-Chat(8*10)             & 55   & 39.5 & 26   & 46.5 & 48.5 & 31.5 & 39   & 37.5 & 35   & 39   \\
LV-Chat(16*10)            & 62   & 41.5 & 27   & 49.5 & 47.5 & 28   & 36   & 38   & 37   & 38   \\
\midrule
\multicolumn{11}{c}{Length 300s}                                                                   \\
\midrule
                          & AS   & AP   & AA   & FA   & UA   & OE   & OI   & OS   & MD   & AL   \\
                          \midrule
VideoChat2(16*1)          & 30.5 & 29   & 63   & 31.5 & 45   & 53   & 39.5 & 32   & 23   & 28.5 \\
VideoChat2(16*10)         & 32   & 28.5 & 40.5 & 24   & 28.5 & 55.5 & 39   & 39   & 19   & 25   \\
VideoChat2(8*10)          & 32   & 28.5 & 40.5 & 24   & 28.5 & 55.5 & 39   & 39   & 19   & 25.5 \\
VideoChatGPT              & 27.5 & 25.5 & 54   & 23.5 & 28   & 53.5 & 26   & 43.5 & 24.5 & 29   \\
VideoLlama               & 25.5 & 23.5 & 41.5 & 26.5 & 38   & 52   & 26   & 33   & 21.5 & 21   \\
LV-Chat(8*10)             & 42.5 & 35.5 & 50   & 26.5 & 36   & 54   & 49.5 & 33.5 & 21.5 & 29   \\
LV-Chat+IFE(8*10)  & 43.5 & 37   & 48.5 & 26.5 & 33.5 & 56   & 50   & 33   & 21   & 29.5 \\
LV-Chat(16*10)            & 41   & 38.5 & 54   & 26.5 & 38.5 & 53.5 & 47   & 32.5 & 20.5 & 28.5 \\
LV-Chat+IFE(16*10) & 42.5 & 37.5 & 54   & 25   & 37   & 53.5 & 52.5 & 32.5 & 20   & 29   \\
\midrule
                          & ST   & AC   & MC   & MA   & SC   & FP   & CO   & EN   & ER   & CI   \\
                          \midrule
VideoChat2(16*1)          & 60   & 44.5 & 28.5 & 58   & 57.5 & 27.5 & 41   & 33   & 35   & 42   \\
VideoChat2(16*10)         & 36.5 & 38.5 & 22.5 & 37   & 58   & 25.5 & 38.5 & 25   & 26   & 39   \\
VideoChat2(8*10)          & 36.5 & 38.5 & 22.5 & 37   & 58   & 25.5 & 38.5 & 25   & 26   & 39   \\
VideoChatGPT              & 38.5 & 29.5 & 23.5 & 28   & 52   & 27   & 38   & 27   & 28.5 & 40.5 \\
VideoLlama               & 30.5 & 29   & 28.5 & 41.5 & 47   & 29   & 33   & 32   & 22.5 & 34.5 \\
LV-Chat(8*10)             & 51.5 & 39   & 25.5 & 45   & 48   & 29.5 & 34.5 & 36.5 & 30   & 34   \\
LV-Chat+IFE(8*10)  & 46   & 40   & 28   & 46   & 48   & 29.5 & 35.5 & 36.5 & 29   & 33   \\
LV-Chat(16*10)            & 49   & 37.5 & 29.5 & 45   & 48.5 & 27   & 34.5 & 36.5 & 35   & 34   \\
LV-Chat+IFE(16*10) & 48.5 & 39   & 29   & 47   & 48.5 & 29.5 & 30   & 35   & 32   & 35   \\
\midrule
\multicolumn{11}{c}{Length 600s}                                                                   \\
\midrule
                          & AS   & AP   & AA   & FA   & UA   & OE   & OI   & OS   & MD   & AL   \\
                          \midrule
VideoChat2(16*1)          & 28.5 & 23   & 63   & 32   & 41.5 & 53   & 39   & 30.5 & 21.5 & 28.5 \\
VideoChat2(16*10)         & 27   & 28   & 39   & 26.5 & 28   & 53   & 35.5 & 39   & 19   & 22.5 \\
VideoChat2(8*10)          & 30   & 28   & 40   & 24.5 & 28.5 & 51   & 35.5 & 39   & 20.5 & 21.5 \\
VideoChatGPT              & 26   & 27   & 56   & 25   & 30   & 52.5 & 26.5 & 40   & 24.5 & 25.5 \\
VideoLlama               & 23.5 & 25   & 40   & 27   & 37.5 & 52.5 & 26   & 33   & 21.5 & 20   \\
LV-Chat(8*10)             & 34   & 32   & 49   & 27.5 & 34.5 & 54   & 49   & 33   & 21.5 & 30   \\
LV-Chat+IFE(8*10)  & 34   & 32   & 49   & 27.5 & 34.5 & 54   & 49   & 33   & 21.5 & 30   \\
LV-Chat(16*10)            & 34.5 & 30.5 & 54   & 24   & 38.5 & 54   & 46   & 33.5 & 19   & 29.5 \\
LV-Chat+IFE(16*10) & 37   & 34   & 50.5 & 24.5 & 38.5 & 53.5 & 48.5 & 32.5 & 19.5 & 28.5 \\
\midrule
                          & ST   & AC   & MC   & MA   & SC   & FP   & CO   & EN   & ER   & CI   \\
                          \midrule
VideoChat2(16*1)          & 51   & 45.5 & 28   & 59.5 & 56.5 & 30.5 & 36.5 & 33   & 32.5 & 43.5 \\
VideoChat2(16*10)         & 38.5 & 38.5 & 22.5 & 36   & 57   & 26   & 39.5 & 25.5 & 25   & 38   \\
VideoChat2(8*10)          & 35.5 & 38.5 & 23   & 33.5 & 59   & 26   & 37.5 & 24.5 & 25   & 36.5 \\
VideoChatGPT              & 38   & 29.5 & 31   & 36.5 & 49   & 25.5 & 38.5 & 28.5 & 26.5 & 39   \\
VideoLlama               & 28   & 29   & 29.5 & 42.5 & 47.5 & 29   & 33   & 31   & 22   & 33.5 \\
LV-Chat(8*10)             & 42.5 & 42.5 & 26   & 43   & 48   & 30   & 33   & 36   & 29.5 & 35.5 \\
LV-Chat+IFE(8*10)  & 42.5 & 42.5 & 26   & 43   & 48   & 30   & 33   & 36   & 29.5 & 35.5 \\
LV-Chat(16*10)            & 44.5 & 37   & 24.5 & 46.5 & 48.5 & 27.5 & 35.5 & 36.5 & 33   & 35   \\
LV-Chat+IFE(16*10) & 47   & 41.5 & 24   & 47   & 47.5 & 27.5 & 37   & 36   & 35   & 33.5\\
\bottomrule
\end{tabular}
\caption{Model performance on extended MVBench}
\label{tab:mvbench_all}
\end{table*}

Here is an example of video s13-d21. The prompt for GPT-4 is:
\begin{lstlisting}
 You are an assistant answering questions based on video contexts. Your answer should be based on the given contexts, but you can also infer the actual video content from the tag information and your common sense. The timed description is a description for the video at the given second. When describing, please mainly refer to the timed description. Don't create a video plot out of nothing.
Contexts for the video: """
Second 9: He took out cutting board
Second 17: He took out knife
Second 22: He took out cucumber
Second 35: He took out plate
Second 47: He washed cucumber
Second 57: Cut off ends of cucumbers
Second 72: He sliced cucumbers
Second 90: He put cucumbers on plate
Second 9: person takes chopping board out
Second 17: person removes knife from draw
Second 22: person removes cucumber out of refrigerator
Second 35: person removes plate out of cabinet
Second 47: person then washes cucumber
Second 57: person then places cucumber on plate
Second 64: perosn then cuts ends off cucumber
Second 72: person then cuts cucumber in slices
Second 90: person then places cucumber on plate.
Second 9: The person gets out a cutting board.
Second 17: The person gets out a knife.
Second 22: The person gets out a cucumber.
Second 35: The person gets out a plate.
Second 47: The person rinses the cucumber.
Second 57: The person cuts the tips off the cucumber.
Second 96: The person slices the cucumber and puts the slices on the plate.
Second 9: The person gets out a cutting board.
Second 17: The person gets out a knife.
Second 25: The person gets out a cucumber.
Second 35: The person gets out a plate.
Second 47: The person rinses the cucumber.
Second 57: The person cuts off the tips  of the cucumber.
Second 72: The person cuts up the cucumber.
Second 90: The person puts the cucumber slices on the plate.
Second 9: The person takes out a cutting board from the drawer.
Second 17: The person takes out a knife from the drawer.
Second 25: The person procures a cucumber from the fridge.
Second 35: The person procures a plate from the cabinet.
Second 47: The person washes the cucumber in the sink.
Second 57: The person cuts the ends off the cucumber then cuts the body into slices.
Second 90: The person sets cucumber slices on the plate.
Second 9: The person takes out a cutting board from the drawer.
Second 17: The person takes out a knife from the drawer.
Second 22: The person procures a cucumber from the fridge then takes a plate from the cabinet.
Second 47: The person washes the cucumber in the sink.
Second 57: The person cuts the ends from the cucumber.
Second 72: The person chops the cucumber into slices on the cutting board.
Second 90: The person sets the cucumber slices on the plate.
Second 9: The person takes out a cutting board from the drawer.
Second 17: The person takes out a knife from the drawer.
Second 22: The person procures a cucumber from the fridge.
Second 35: The person procures a plate from the cabinet.
Second 47: The person washes the cucumber in the sink.
Second 57: The person cuts the ends off the cucumber.
Second 72: The person slices the cucumber on the cutting board.
Second 90: The person sets the sliced cucumber on the plate.
Second 9: He goes to the drawer and takes out a cutting board and knife.
Second 25: He goes to the refrigerator and takes out a cucumber.
Second 35: He goes to the cupboard and takes out a plate and places it on the counter.
Second 50: He goes to the sink and washes the cucumber.
Second 57: He then cuts off the ends of the cucumber and then slices the cucumber.
Second 72: He picks up the cucumber and places it on the plate.
Second 9: He opens the drawers and takes out a cutting board and a knife.
Second 25: He gets a cucumber from the refrigerator and a plate from the cabinet.
Second 47: He sets the plate down and washes the cucumber in the sink.
Second 57: He puts the cucumber on the plate and dries off his hands.
Second 64: He uses the knife to cut off the ends of the cucumbers.
Second 72: He uses the knife to slice the cucumber into smaller pieces.
Second 96: He picks up the pieces of cucumber and places them on the plate.
Second 9: The person takes out a cutting board from the drawer.
Second 17: The person takes out a knife from the drawer.
Second 22: The person procures a cucumber from the fridge.
Second 35: The person procures a plate from the cabinet.
Second 47: The person washes the cucumber in the sink.
Second 57: The person chops the ends off the cucumber on the cutting board.
Second 72: The person slices the cucumber on the cutting board.
Second 90:  The person sets the sliced cucumber on the plate.
Second 9: He gets out the cutting board, knife, plate, and cucumber from drawers and the refrigerator.
Second 50: He washes the cucumber in the sink and puts it on the plate.
Second 60: He wipes his hand on the rag.
Second 65: He chops the ends off the cucumber.
Second 72: He chops the cucumber into pieces with the knife.
Second 96: He gathers the chopped cucumbers together on a plate.
Second 9: the man is opening the drawer
Second 17: the man is placing a knife on the cutting board he got from the drawer
Second 22: the man is opening the refridgerator
Second 25: he got a cucumber from the refridgerator
Second 35: the man is washing the cucumber
Second 50: the man is slicing the cucumber width-wise
Second 96: then the man arranges the cucumbers on a plate
Second 9: The man places a cutting board on the counter.
Second 17: The man places a knife on the cutting board.
Second 22: The man gets a cucumber out of the refrigerator.
Second 35: The man gets a plate out of the cupboard.
Second 47: The man rinses the cucumber.
Second 57: The man dries his hands.
Second 64: The man cuts the ends off of the cucumber.
Second 72: The man cuts the cucumber into slices.
Second 96: The man arranges the cucumber slices on the plate.
Second 9: The person takes a cutting board from the drawer.
Second 17: The person removes a large knife from a drawer and puts it on the cutting board.
Second 25: The person takes a cucumber from the fridge.
Second 35: The person takes a plate from the cabinet.
Second 47: The person washes the cucumber in the sink.
Second 57: The person dries his hands with the towel.
Second 64: The person slices the cucumber.
Second 90: The person places the sliced cucumber onto the plate.
Second 9: He places the cutting board on the counter
Second 17: He placed the knife on the cutting board
Second 25: He got a cucumber from the refrigerator
Second 35: The man washed the cucumber in the sink
Second 57: The man dried his hands on the towel
Second 64: The man cut the cucumber into slices
Second 90: The man placed the cucumber slices on the plate
"""
Question:Could you please describe what is happening in the video?
\end{lstlisting}
And the response is:
\begin{lstlisting}
The video is showing a person preparing a cucumber. They start by taking out a cutting board and a knife from a drawer. Then, they get a cucumber from the refrigerator and a plate from the cabinet. After that, they wash the cucumber in the sink. They then cut off the ends of the cucumber and slice it into pieces on the cutting board. Finally, they place the sliced cucumber on the plate.
\end{lstlisting}
We manually inspect the generated responses and query GPT-4 again if there are unreasonable responses or clear hallucinations. 

\subsection{Prompt for different models}
\label{sub:prompt}
\subsubsection{System Prompt}
For Video-ChatGPT, we use the system prompt (as used in the original paper):
\begin{lstlisting}
You are Video-ChatGPT, a large vision-language assistant. You are able to understand the video content that the user provides, and assist the user with a variety of tasks using natural language. Follow the instructions carefully and explain your answers in detail based on the provided video.   
\end{lstlisting}

For VideoChat2, Video-Llama, and our own model, we use the same system prompt from MVBench\cite{mvbench}:
\begin{lstlisting}
Carefully watch the video and pay attention to the cause and sequence of events, the detail and movement of objects, and the action and pose of persons.   
\end{lstlisting}

\subsubsection{Dataset-specific prompt}
In TaCoS generation, the user asks the assistant: 
\begin{lstlisting}
Based on your observations, describe what is happening in the video as detailed as possible.
\end{lstlisting}
In QA datasets (MVBench and EgoSchema), we use the same format as in MVBench. Following is an example:
\begin{lstlisting}
Question: What happened after the person took the food?
Options:
(A) Ate the medicine.
(B) Tidied up the blanket.
(C) Put down the cup/glass/bottle.
(D) Took the box.
Only give the best option. 
\end{lstlisting}

\section{Detailed Results}
\subsection{Model performance on all subsets of MVBench}
\label{sub:all_subsets}
Table~\ref{tab:mvbench_original} shows the results on the original MVBench and Table~\ref{tab:mvbench_all} shows the results on the augmented MVBench with Street-Scene.

\subsection{Detailed Results for the Effectiveness of IFE}
\label{sub:detailed_results_for_the_effectiveness_of_ife}
Table~\ref{tab:acc_join1_600} shows the performance of \ours and VideoChat2 on the 4 chosen subsets extended to different lengths. 
\begin{table*}[ht]
  \centering
    \begin{tabular}{l|rrrr|r}
    \toprule
          & AS & \multicolumn{1}{l}{AP} & \multicolumn{1}{l}{UA} & OI & Avg \\
      \midrule
    \multicolumn{6}{c}{Length 100s} \\
    \midrule
    VideoChat2 & 38.5  & 33    & 46.5  & 57.5  & 43.875 \\
    LV-Chat(16*6)  & 54    & 42    & 48    & 65.5  & 52.375 \\
    LV-Chat(16*6+IFE)  & 51.5  & 44    & 58.5  & 64.5  & 54.625 \\
    \midrule
    \multicolumn{6}{c}{Length 200s} \\
    \midrule
    VideoChat2 & 35    & 29.5  & 44.5  & 47    & 39 \\
    LV-Chat(16*6)  & 44.5  & 42.5  & 48    & 60    & 48.75 \\
   LV-Chat(16*6+IFE) & 47    & 43    & 47    & 60    & 49.25 \\
   \midrule
    \multicolumn{6}{c}{Length 300s} \\
    \midrule
    VideoChat2 & 30.5  & 29    & 45    & 39.5  & 36 \\
    LV-Chat(16*6)  & 46    & 34    & 41.5  & 56    & 44.375 \\
    LV-Chat(16*6+IFE) & 46    & 39.5  & 42    & 56.5  & 46 \\
    \midrule
    \multicolumn{6}{c}{Length 400s} \\
    \midrule
    VideoChat2 & 30    & 26.5  & 44.5  & 40    & 35.25 \\
    LV-Chat(16*6)  & 38    & 36.5  & 40    & 50.5  & 41.25 \\
    LV-Chat(16*6+IFE) & 40    & 38.5  & 40.5  & 54    & 43.25 \\
    \midrule
    \multicolumn{6}{c}{Length 500s} \\
    \midrule
    VideoChat2 & 28.5  & 27.5  & 46    & 36.5  & 34.625 \\
    LV-Chat(16*6)  & 38.5  & 38    & 41.5  & 48    & 41.5 \\
    LV-Chat(16*6+IFE) & 40.5  & 36    & 39    & 49    & 41.125 \\
    \midrule
    \multicolumn{6}{c}{Length 600s} \\
    \midrule
    VideoChat2 & 28.5  & 23    & 41.5  & 39    & 33 \\
   LV-Chat(16*6)  & 37.5  & 35    & 43    & 47.5  & 40.75 \\
    LV-Chat(16*6+IFE) & 43.5  & 34.5  & 41    & 48.5  & 41.875 \\
    \bottomrule
    \end{tabular}%
      \caption{Accuracy on different video length}
  \label{tab:acc_join1_600}
\end{table*}%

% Table generated by Excel2LaTeX from sheet 'acc_vs_video_length'

\section{Generation Cases}
\label{sec:gen_cases}
Some other TACoS generation cases are shown in Table ~\ref{tab:append_tacos_cases}. The Reference is the summary of the captions generated by OpenAI's GPT4.
\begin{table*}[]
\footnotesize
    \centering
    \begin{tabular}{c|p{5.5cm}|p{4cm}|p{2.5cm}}
    \toprule
        Video & Reference & \ours & VideoChat2\\
        \midrule
        s14-d46 & The video is a tutorial on how to properly cut and prepare a pineapple. The woman starts by gathering all the necessary tools and the pineapple. She uses two different knives, a cutting board, a bowl, and a storage container. She begins by cutting off the top and bottom of the pineapple, then proceeds to cut off the sides using a smaller knife. She makes sure to remove all the hard spots on the pineapple's surface. After the pineapple is fully peeled, she slices it into rounds and places them in a stainless steel bowl. After she's done cutting, she cleans her knives and disposes of the pineapple scraps.
& The video shows a woman standing in a kitchen cutting a pineapple on a cutting board. She is wearing a black shirt and a black apron. The kitchen has a black counter and a black sink. The woman is cutting the pineapple into slices with a knife. She is standing in front of a black stove and a black microwave. The video is shot from an overhead angle, giving a bird's eye view of the kitchen. The woman is cutting ...
& The video shows a woman cutting a pineapple on a cutting board.
\\ \hline
s36-d27 
& The video is a tutorial on how to properly cut and prepare an onion. The person in the video starts by taking out an onion, a cutting board, and a knife. They then cut off the ends of the onion and peel off the skin. After that, they cut the onion in half and peel off any remaining skin. The person then proceeds to dice the onion into small pieces. After the onion is diced, they get out a plate and put the diced onion on it. Finally, they add some lemon juice to the diced onion.
& In the video, we see a man standing in a kitchen, looking at a plate of food. He then takes a knife and begins cutting the food on the plate. He continues to cut the food, and we see him cutting the food into smaller pieces. He then puts the food on a plate and puts it on the counter. He then takes a knife and begins cutting the food again. He continues to cut the food, and we see him cutting the food into smaller pieces...
& The video shows a man cutting onions on a cutting board. 
\\  \hline
s36-d23
& The video is a cooking tutorial where a man is preparing a carrot dish. He starts by washing his hands and gathering the necessary ingredients and tools, including carrots, a bowl, a cutting board, and a knife. He washes the carrots, then proceeds to chop them into small pieces. After chopping, he washes his cutting board and knife. He then turns on the stove, takes out a pan, and adds oil to it. He adds the chopped carrots to the pan and seasons them with various spices. He continues to cook the carrots, stirring them occasionally. Once the carrots are cooked, he transfers them to a bowl. Finally, he cleans up his workspace, washing and putting away the used utensils and equipment.
& The video shows a man standing in a kitchen and preparing food. He is wearing a black shirt and a black and white checkered shirt. He is cutting vegetables and placing them in a pan. He then puts the pan on the stove and starts cooking. The video also shows a man standing in the kitchen and cutting vegetables. He is wearing a black shirt and a black and white checkered shirt. He is cutting vegetables and placing them ...
& The video shows a man in a kitchen cutting vegetables and frying them in a pan.
\\  \hline
s13-d21
& The video is showing a person preparing a cucumber. They start by taking out a cutting board and a knife from a drawer. Then, they get a cucumber from the refrigerator and a plate from the cabinet. After that, they wash the cucumber in the sink. They then cut off the ends of the cucumber and slice it into pieces on the cutting board. Finally, they place the sliced cucumber on the plate.
& The video shows a man standing in a kitchen, cutting vegetables on a cutting board. He then places the vegetables on a plate and puts it on a counter. The man then takes a knife and cuts the vegetables. He then puts the vegetables on a cutting board and cuts them again. He then puts the vegetables on a plate and puts it on the counter. The man then takes a knife and cuts the vegetables again. He then puts the vegetables ...
& The video shows a man cutting vegetables on a cutting board.
\\

\bottomrule
    \end{tabular}
    \caption{TACoS generation cases}
    \label{tab:append_tacos_cases}
\end{table*}

\end{document}